\documentclass[sigconf]{acmart}
\AtBeginDocument{%
  }

\setcopyright{acmlicensed}
\copyrightyear{2018}
\acmYear{2018}
\acmDOI{XXXXXXX.XXXXXXX}
\acmISBN{978-1-4503-XXXX-X/2018/06}

\usepackage{multirow}
\usepackage{float}
\usepackage{stfloats}
\usepackage{fancyhdr}
\acmSubmissionID{500}

\renewcommand\footnotetextcopyrightpermission[1]{}
\begin{document}
\newcommand{\model}{HM-RAG}
\title{\model: Hierarchical Multi-Agent Multimodal Retrieval Augmented Generation}


\author{Pei Liu$^{1,2}$, Xin Liu$^{2}$, Ruoyu Yao$^{2}$, Junming Liu$^{1}$, Siyuan Meng$^{1}$, Ding Wang$^{1*}$, Jun Ma$^{23*}$}
\affiliation{%
  \institution{$^{1}$Shanghai Artificial Intelligence Laboratory \quad $^{2}$The Hong Kong University of Science and Technology (Guangzhou) \quad}
  \city{$^{3}$The Hong Kong University of Science and Technology} \\
  \country{pliu061@connect.hkust-gz.edu.cn\quad wangding@pjlab.org.cn \quad jun.ma@ust.hk}
  }







\renewcommand{\shortauthors}{}

\begin{abstract}

While Retrieval-Augmented Generation (RAG) augments Large Language Models (LLMs) with external knowledge, conventional single-agent RAG remains fundamentally limited in resolving complex queries demanding coordinated reasoning across heterogeneous data ecosystems. We present {\model}, a novel Hierarchical Multi-agent Multimodal RAG framework that pioneers collaborative intelligence for dynamic knowledge synthesis across structured, unstructured, and graph-based data. The framework is composed of three-tiered architecture with specialized agents: a Decomposition Agent that dissects complex queries into contextually coherent sub-tasks via semantic-aware query rewriting and schema-guided context augmentation; Multi-source Retrieval Agents that carry out parallel, modality-specific retrieval using plug-and-play modules designed for vector, graph, and web-based databases; and a Decision Agent that uses consistency voting to integrate multi-source answers and resolve discrepancies in retrieval results through Expert Model Refinement. This architecture attains comprehensive query understanding by combining textual, graph-relational, and web-derived evidence, resulting in a remarkable 12.95\% improvement in answer accuracy and a 3.56\% boost in question classification accuracy over baseline RAG systems on the ScienceQA and CrisisMMD benchmarks. Notably, HM-RAG establishes state-of-the-art results in zero-shot settings on both datasets. Its modular architecture ensures seamless integration of new data modalities while maintaining strict data governance, marking a significant advancement in addressing the critical challenges of multimodal reasoning and knowledge synthesis in RAG systems. Code is available at \url{https://github.com/ocean-luna/HMRAG}.
\end{abstract}



\keywords{Retrieval-Augmented Generation (RAG), Multimodal Representation, Multi-agent Systems, Multi-source RAG}


\maketitle
\section{Introduction}
\label{sec:intro}


\begin{figure}
    \centering
    \includegraphics[width=1.0\linewidth]{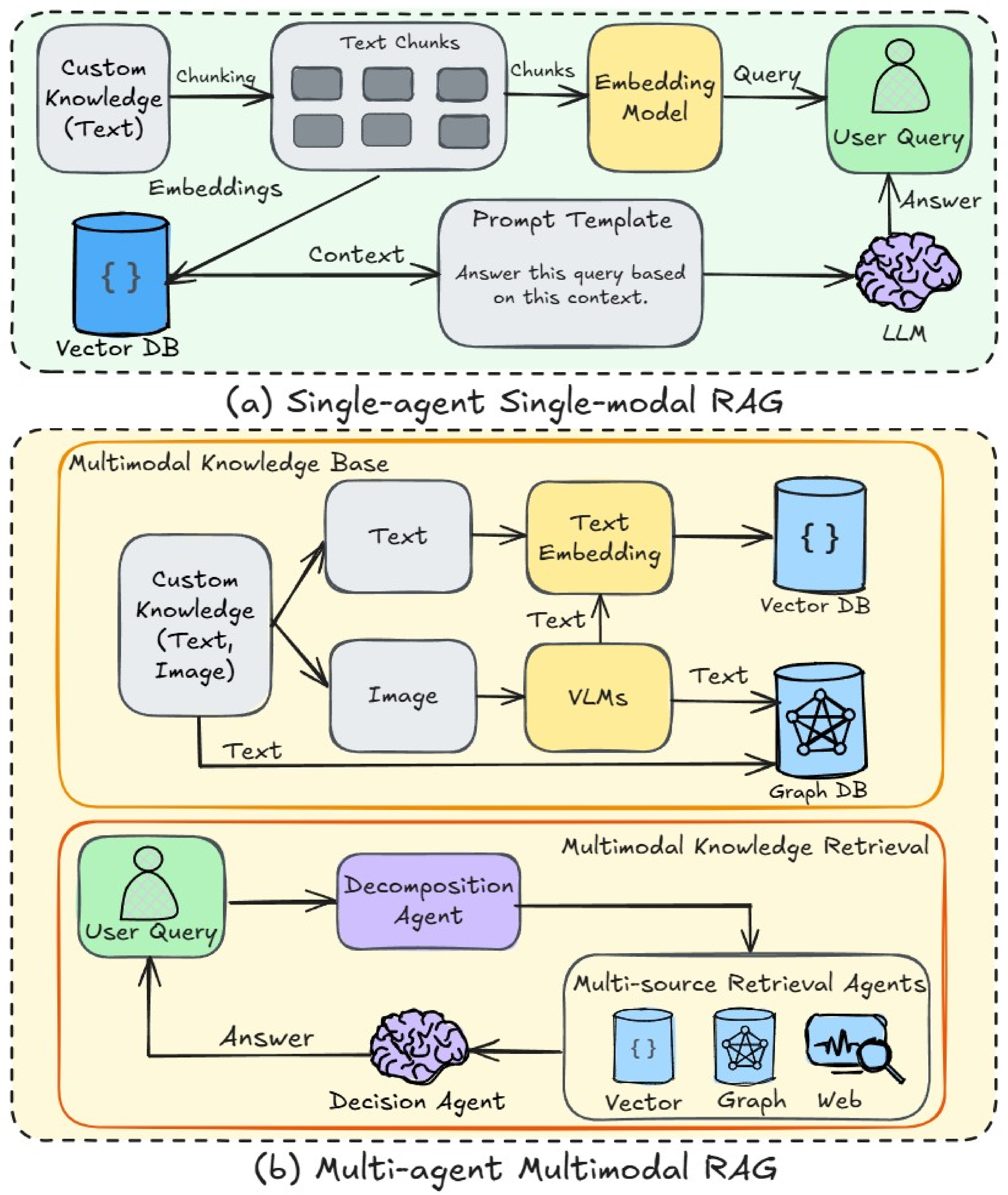}
    \caption{Comparison of ($a$) single-agent single-modal RAG and ($b$) multi-agent multimodal RAG. The multi-agent multimodal RAG processes multimodal data by converting them into vector and graph databases. It leverages multi-source retrieval across vector, graph, and web-based databases, enabling more comprehensive and efficient information retrieval. This advanced approach allows the multi-agent multimodal RAG to achieve superior performance in handling complex queries and diverse data types, setting it apart from the more limited single-agent single-modal RAG.}
    \label{fig:enter-label}
\end{figure}


In an era defined by the rapid proliferation of data, the ability to efficiently retrieve relevant information from heterogeneous sources has emerged as a fundamental pillar of modern information systems \cite{fey2023relational}. Multimodal retrieval systems, which integrate text, images, vectorized data, and web-based content, are becoming indispensable across domains such as e-commerce, healthcare, and scientific research \cite{zhong2023knowledge}. These systems enable the seamless navigation of diverse data types, empowering users to derive actionable insights across multiple modalities. However, despite remarkable progress in recent years, multimodal retrieval continues to present significant challenges. The complexity arises from the need to reconcile the diversity of query types, the heterogeneity of data formats, and the varying objectives of retrieval tasks, all of which demand sophisticated solutions to bridge the gap between data representation and user intent.

The evolution of retrieval technologies has historically centered on single-modal architectures, where queries and retrieval mechanisms operate within a single predefined modality \cite{lewis2020retrieval, anand2023context}. While text-based retrieval-augmented generation (RAG) systems have demonstrated robust performance in processing linguistic information \cite{naveed2023comprehensive}, their inability to handle visual content has spurred the development of image-based RAG approaches \cite{hu2024mplug, chen2024mllm, luo2024layoutllm}. However, current multimodal implementations face a critical bottleneck: Although image-based RAG systems excel at visual content processing, they often fail to establish coherent cross-modal correlations between visual elements and textual context. This limitation is particularly acute in multimodal question answering, where systems must integrate visual perception with textual semantics to generate contextually relevant responses.

Recently, graph-based retrieval frameworks have been proposed to enhance the modeling of textual interdependencies based on the construction of knowledge graphs, represented by GraphRAG \cite{edge2024local} and LightRAG \cite{guo2024lightrag}. These approaches are further extended to processing multimodal inputs \cite{liu2025aligning}, where graph structures are leveraged for the accurate capture of cross-modal relationships.
Despite these advances, graph-based methods face an inherent trade-off: while they effectively capture high-level modality interactions, they often sacrifice fine-grained information fidelity. This becomes problematic in scenarios requiring precise textual segment retrieval, as the abstraction process inherent to graph modeling obscures granular textual details critical for nuanced analysis.

Meanwhile, another critical challenge has been noticed in reconciling the complementary strengths of
different modalities \cite{gao2023retrieval, khattab2020colbert, faysse2024colpali}.
Textual modalities excel at encoding granular semantic details and conceptual relationships, while visual modalities, by contrast, are capable of capturing spatial context and facilitating spatial relationship understanding. Current modality-specific systems \cite{lewis2020retrieval, xia2024mmed} exhibit critical limitations in cross-modal synthesis, producing retrieval outcomes that are either overspecialized in textual precision or confined to visual pattern recognition. This modality isolation creates systemic vulnerabilities in heterogeneous data environments, where the absence of cross-modal alignment protocols risks critical information loss during retrieval operations. For instance, visual queries in text-centric systems fail to map conceptual questions to illustrative elements, while text-intensive inquiries in vision-oriented frameworks lack mechanisms for lexical disambiguation. These architectural gaps highlight the urgent need for frameworks that can harmonize granular semantic detail with cross-modal contextual coherence.

To address these challenges, we introduce Hierarchical Multi-Agent Retrieval-Augmented Generation ({\model}), a novel framework that enhances multimodal retrieval through coordinated multi-agent collaboration. {\model} employs a three-tiered architecture with specialized agents operating in the RAG pipelines. The Decomposition Agent analyzes query intent and dynamically rewrites requests to ensure cross-modal compatibility. The Multi-Source Retrieval Agent conducts parallel knowledge acquisition via lightweight multimodal retrievals across diverse data sources, including vectors, graphs, and web-based databases. Finally, the Decision Agent synthesizes and refines candidate responses using domain-specific verification strategies to ensure accuracy and coherence. This hierarchical design systematically orchestrates text-image evidence integration through structured agent interactions, enabling layered reasoning. Unlike conventional approaches, {\model} combines query decomposition, parallelized information retrieval, and expert-guided answer refinement to achieve efficient and contextually relevant responses.
Our contributions are summarized as follows:
\begin{itemize}
    \item We propose a novel \textbf{Modularized Hierarchical Framework} that modularizes query processing into specialized agent-based components, and this facilitates scalable and efficient multimodal retrieval.
    \item We enable \textbf{Multi-source Plug-and-play Retrieval Integration}, which offers seamless connectivity across diverse data sources. By efficiently routing queries to vector, graph, and web-based retrieval agents, our approach ensures flexibility and efficiency in handling heterogeneous data environments, streamlining complex information retrieval processes.
    \item We employ \textbf{Expert-guided Refinement} processes to enhance response quality to ensure both operational efficiency and contextual precision through minimal expert oversight.
    \item We demonstrate the effectiveness of {\model} through extensive experiments on benchmark datasets, and the results attain \textbf{State-of-the-art Performance} on the ScienceQA and CrisisMMD benchmarks.
\end{itemize}

\section{Related Work}
\label{sec:formatting}

\subsection{Retrieval-Augmented Generation}

RAG systems have evolved significantly to enhance their multimodal reasoning capabilities \cite{lewis2020retrieval, guu2020retrieval, genesis2025integrating, csakar2025maximizing}. Initially, text-based RAG systems integrated Large Language Models (LLMs) with external textual knowledge, improving performance in question answering by retrieving relevant text fragments \cite{izacard2022few, asai2023self, zhang2024raft}. However, as visually rich documents became more prevalent, the limitations of text-only systems became evident, prompting the development of image-based RAG approaches \cite{bag2024rag, bonomo2025visual, riedler2024beyond, liu2025siqa}. While these methods aimed to retrieve visual content for Large Vision-Language Models (VLMs), they faced challenges in effectively integrating text and image modalities, as the retrieval processes were largely independent, hindering a deep understanding of their interrelationships.

To address these challenges, graph-based RAG systems emerged, leveraging structured knowledge representations to capture both inter-modal and intra-modal semantic relationships \cite{dong2024advanced, jeong2024graph, guo2024lightrag, procko2024graph}. These systems utilize vector-space embeddings and topological relationships to model complex document structures, enabling the retrieval of semantically coherent contexts that go beyond simple text fragments \cite{wu2024medical, edge2024local, mavromatis2024gnn}. Graph-based RAG systems are particularly effective in understanding relationships between text and images, as well as extracting relationships within the text itself \cite{liu2025aligning}. 
However, current RAG implementations often rely on single-source retrieval, limiting their ability to handle complex queries that require simultaneous processing of vector, graph, and web-based databases \cite{gupta2024comprehensive}. This limitation is particularly significant in applications requiring private data retrieval and real-time updates, where the absence of integrated multi-source retrieval capabilities can lead to incomplete or outdated information. To fully leverage the strengths of each data modality and meet the demands of dynamic and heterogeneous data environments, RAG systems must evolve to support coordinated multi-source retrieval and synthesis.

\subsection{Agents in RAG}

RAG has become a key paradigm for knowledge-intensive tasks by integrating retrieval mechanisms with generative models, significantly enhancing language model capabilities. However, traditional RAG implementations often rely on static pipelines that struggle with multimodal query processing \cite{schick2023toolformer, cheng2024exploring}. Recent agent-based RAG architectures have addressed these limitations by improving system modularity and operational flexibility \cite{han2025mdocagent, jeong2024study, e2025rag}.
The agent-oriented approach breaks down query processing into specialized components like semantic parsing, cross-modal retrieval, and context-aware generation, allowing targeted optimization while maintaining overall adaptability. PaperQA \cite{lala2023paperqa} exemplifies this by leveraging academic literature to generate evidence-based responses, reducing hallucinations in scientific applications.

Building on this, Active RAG methodologies like FLARE \cite{jiang2023active} introduce temporal dynamism through anticipatory retrieval, enhancing performance in extended text generation. Despite these advances, challenges in multimodal integration persist. Emerging Dynamic RAG approaches \cite{su2024dragin, toro2024dynamic} propose entity-aware augmentation strategies to dynamically incorporate retrieved entity representations, addressing context window limitations while preserving semantic coherence.
Our {\model} framework synthesizes these innovations through a hierarchical multi-agent architecture leveraging LLMs' semantic comprehension. This design enables dynamic query adaptation and multimodal retrieval, providing an optimized solution for complex information retrieval and generation tasks across diverse data modalities. By integrating these advancements, {\model} addresses key challenges in multimodal reasoning and knowledge synthesis, paving the way for more robust and adaptable RAG systems.

\section{Methodology}

\begin{figure*}
    \centering
    \includegraphics[width=1.0\linewidth]{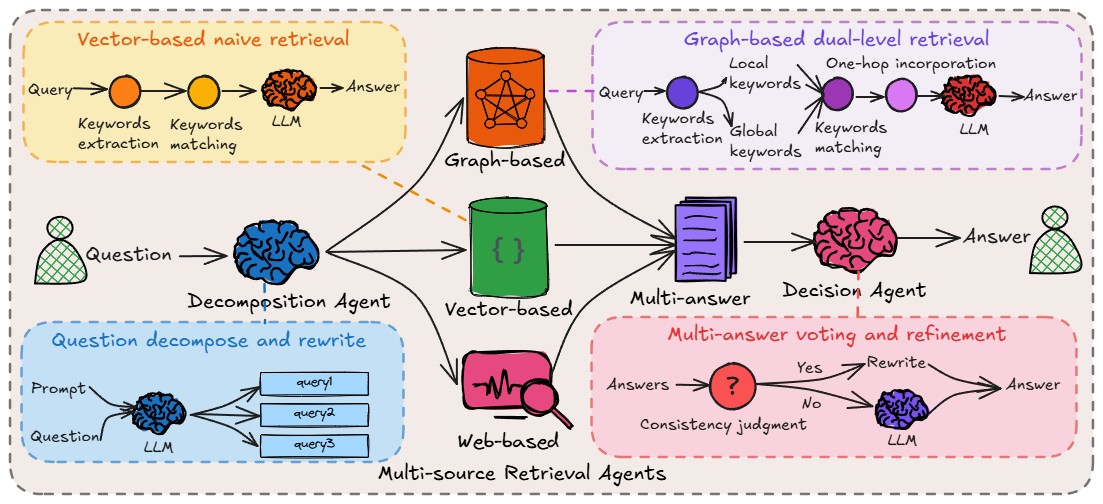}
    \caption{Overview of {\model}. A multi-agent multi-modal framework operates in three stages: First, the Decomposition Agent uses an LLM to rewrite and decompose the question into several sub-queries. Second, the Multi-source Retrieval Agent retrieves the top-k relevant documents from vector-, graph- and web-based sources as needed. Finally, the Decision Agent provides a voting mechanism and refinement process to generate the final answer.}
    \label{fig:architecture}
\end{figure*}

We introduce {\model}, a novel framework tackling complex challenges in RAG systems. As depicted in Figure \ref{fig:architecture}, {\model} features an innovative multi-agent, multimodal architecture with specialized agents for information extraction and multi-source retrieval. Given a natural language question $q$ and a reference document $\mathcal{D}$, RAG retrieves semantically relevant content from $\mathcal{D}$, integrating it with generative language models to produce answers strictly grounded in $\mathcal{D}$. This approach advances multimodal question answering and multi-agent RAG capabilities.
The subsequent sections provide a detailed exposition of {\model}'s architectural design. Through this systematic description, we elucidate the framework's core mechanisms for effectively integrating and utilizing multimodal information and multi-source retrieval, ultimately leading to enhanced accuracy in RAG applications.


\subsection{Multimodal Knowledge Pre-Processing}
This section focuses on multimodal data processing, aiming to convert textual data and visual images into vector and graph database representations for enhanced retrieval operations. Our methodology employs VLMs to transcode visual information into textual representations, which are subsequently integrated with original text corpora to jointly construct vector and graph databases.

\subsubsection{Multimodal Textual Knowledge Generation}
Conventional entity-centric approaches for multimodal knowledge extraction rely on predefined categorical boundaries, limiting their capacity to recognize novel visual concepts. We utilize the BLIP-2's framework \cite{li2023blip} to harness the open vocabulary potential of pretrained VLMs. Building upon the generalized vision to language conversion paradigm:

\begin{equation}
    T_{v} = {\mathcal{D}_{blip2}}({f_{align}}({\mathcal{E}_{blip2}}(I_v)))
\end{equation}
where visual encoder $\mathcal{E}_{blip2}$ extracts features from input image $I_v$ and cross-modal alignment module $f_{align}$ bridges vision-language semantics. Our framework addresses the critical limitation of oversimplified machine-generated descriptions, particularly addressing BLIP-2's over-condensed outputs that lack visual specificity, through contextual refinement mechanisms leveraging original textual data. 

This process is divided into three synergistic phases. \textbf{Hierarchical visual encoding} via established architectures \cite{he2016deep, liu2021swin, dosovitskiy2020image} to generate patch embeddings $V_i \in \mathcal{R}^{d_v\times N_p}$. \textbf{Cross-modal interaction} where learnable queries $Q_i \in \mathcal{R}^{d_q\times L_q}$ attend to visual features through scaled dot product attention, dynamically weighting spatial semantic correlations. \textbf{Context-aware text generation} that fuses latent text features from prior descriptions $T_v^{i,t}$ with cross-modal representations for autoregressive decoding. Contextual refinement during this phase enhances semantic alignment, achieving measurable reductions in descriptive ambiguity and lexical sparsity for the final output $T_v$.

The resultant multimodal textual knowledge base is subsequently formed through the systematic integration of original textual inputs with generated textualizations.

\begin{equation}
    T_{m} = Concate(T, T_{v})
\end{equation}
where $T$ corresponds to the source textual corpus and $T_m$ represents the multimodal textual aggregation formed through heterogeneous fusion processes.

\subsubsection{Multimodal Knowledge Graphs Construction}
We establish multimodal knowledge graphs (MMKGs) by synergizing VLM-enhanced descriptions with LLM-based structural reasoning. Building upon the refined visual descriptions $T_v$ generated by VLMs, optionally fused with external textual knowledge $T$, we employ the LightRAG framework \cite{guo2024lightrag} for efficient multi-hop reasoning and dynamic knowledge integration:
\begin{equation}
    G = LightRAG(T_v, T)
\end{equation}
LightRAG processes multimodal inputs through a hybrid extraction strategy. \textbf{Entity-Relation Extraction}: a specialized function $f$ decomposes inputs into entities $E=\{e_1, \cdots, e_n\}$ and relation triplets $R=\{(h_i,r_i,t_i)\}$, where $h, t \in E$ represent head/tail entities and $r \in R$ denotes relations. \textbf{Dual-level Reasoning Augmentation:} Dual-scale retrieval mechanisms $Retrieve_{global+local}$ dynamically fetch relevant triplets during inference; global retrieval identifies thematic clusters while local extraction focuses on entity-specific connections.

The constructed MMKG $G=(E,R)$ formalizes knowledge as triplets 
$(h,r,t)$, where entities encompass both visual concepts from 
$T_v$ and textual knowledge from $T$. Crucially, visual data storage locations are embedded during graph construction, enabling cross-modal grounding. This architecture establishes a bidirectional knowledge enhancement framework: language models achieve visual-contextualized reasoning through visual-semantic relationships embedded in $G$, and vision-language models dynamically update knowledge embeddings via continuous multimodal integration, effectively mitigating hallucination probabilities through representation consistency constraints.

\subsection{Decomposition Agent for Multi-intent Queries}
The Decomposition Agent is a pivotal component of the proposed framework, designed to break down complex, multi-intent user queries into coherent and executable sub-tasks. This agent addresses a critical limitation of traditional systems, which often struggle to process compound queries requiring joint reasoning across multiple data sources. By leveraging a hierarchical parsing mechanism, the Decomposition Agent identifies the underlying structure of user queries and decomposes them into atomic units, with each targeting a specific data modality or retrieval task.

The proposed framework operates in two stages, both driven by task-specific LLM-prompting strategies.
\textbf{Decomposition Necessity Judgment.} The agent first determines whether the input question $Q$ contains multiple intents using a binary decision prompt that instructs the LLM to classify it as single-intent or multi-intent. If the output is multi-intent, $Q$ proceeds to decomposition. Otherwise, return question $Q$ directly. \textbf{Intent Decomposition.} The LLM decomposes 
$Q$ into candidate sub-questions $q = \{q_1, \cdots, q_n\}$ using a structured prompt: "Decompose the reasoning steps of the original question into 2 to 3 simply and logically connected sub-questions based on its intent while retaining keywords from the original question." inspired by \cite{li2025topology}.

\subsection{Multi-source Plug-and-Play Retrieval Agents}

We propose a modular multi-agent retrieval framework that dynamically composes heterogeneous multimodal search strategies through standardized interfaces. By decoupling retrieval functionalities into three specialized agents—vector-based retrieval agent, graph-based retrieval agent, and web-based retrieval agent—the system achieves domain-agnostic adaptability while ensuring interoperability across diverse search scenarios. Each agent adheres to unified communication protocols, enabling seamless integration of vector semantic search, graph topological exploration, and real-time web retrieval capabilities. This design allows each retrieval agent to function as a plug-and-play component, ensuring that they can be easily integrated or replaced without affecting the overall system performance. This modularity not only enhances flexibility but also maintains task-specific optimization objectives, making the framework highly adaptable to various applications and data modalities.

\subsubsection{Vector-based Retrieval Agent for Fine-Grained Information}
 This agent leverages a naive retrieval architecture \cite{guo2024lightrag} to search unstructured textual corpora efficiently. 
Given the user query $q$, the system first computes its semantic embedding $h_q$ using an encoder $\mathcal{E}_{text}$:
\begin{equation}
    h_q = \mathcal{E}_{text}(q)
\end{equation}
where $h_q \in \mathbb{R}^d$ represents the query's embedding in a $d$-dimensional vector space.

Next, the system computes the semantic similarity between the query embedding $h_q$ and all document embeddings $h_j$ using cosine similarity:

\begin{equation}
    s_j = \frac{h_q^Th_j}{||h_q||||h_j||}, \quad {\forall}j \in [1, M]
\end{equation}
where $j \in [1,M]$, with $M$ being the total number of documents. The similarity score $s_j$ quantifies how closely each document aligns with the query, forming the basis for ranking retrieved documents.

Based on the similarity scores, the system retrieves the top-$k$ most relevant documents:
\begin{equation}
    \mathcal{R}_k = \{c_1, \cdots,c_k\} \quad s.t. \quad s_1 \geq s_2 \geq \cdots \geq s_k
\end{equation}
where $\mathcal{R}_k$ denotes the set of top-$k$ retrieved contexts, ensuring that only the most relevant information is used for subsequent processing.

Subsequently, the language model generates answers $\mathcal{A}_v$ conditioned on retrieved contexts through constrained decoding:
\begin{equation}
   \mathcal{A}_v = \mathcal{P}(q,\mathcal{R}_k) = Concate(q, Context, \{c_1, \cdots,c_k\})
\end{equation}
where $\mathcal{P}$ represents the generation process, which concatenates the query $q$, retrieved contexts $\{c_1, \cdots,c_k\}$, and additional contextual information to produce the final answer.

Specifically, the conditional probability of generating a token sequence y given the query q and retrieved contexts $\mathcal{R}_k$ is modeled as:
\begin{equation}
    p(y|q, \mathcal{R}_K) = \prod_{t=1}^Tp_{lm}(y_t|y_{<t},q, \mathcal{R}_K)
\end{equation}
where $p_{lm}$ denotes the conditional probability of a token in the auto-regressive generation process of a language model, ensuring that the generated answer is contextually coherent.

Furthermore, the attention mechanism explicitly incorporates retrieved content into the generation process:
\begin{equation}
    Attention(Q,K,V)=softmax(\frac{Q[h_q;H_{\mathcal{R}}]^T}{\sqrt{d_k}})[h_q;H_{\mathcal{R}}]
\end{equation}
where $H_{\mathcal{R}} \in \mathbb{R}^{K \times d}$ stacks the embeddings of retrieved chunks, and $[h_q;H_{\mathcal{R}}]$ concatenates the query embedding with the retrieved chunk embeddings, enhancing the model's ability to focus on relevant information. To ensure the reliability of the generated answers, constraints enforce top-$p=1.0$ and a temperature of 0, ensuring deterministic decoding based on the highest probability tokens. This minimizes the risk of hallucination and ensures factual accuracy.

\subsubsection{Graph-based Retrieval Agent for Relational Information}

This agent leverages LightRAG's graph traversal capabilities to resolve multi-hop semantic queries over MMKGs \cite{guo2024lightrag}. Given an input query $q$, the agent constructs a context-aware subgraph $G_q \subseteq G$ by dynamically retrieving entities and relations through the joint attention mechanism of LightRAG. The subgraph is defined as:

\begin{equation}
    G_q = \{(h,r,t)|LightRAG_{graph}(q,h,r,t) > \tau\}
\end{equation}
where $LightRAG_{graph}$ computes relevance scores by aligning query embeddings with graph triplet representations through cross-modal attention, ensuring that only highly relevant triplets are included in the subgraph. 

To efficiently address complex queries, the agent employs a hierarchical search strategy that balances efficiency and comprehensiveness. First, the agent prioritizes local 1-hop neighbors of query-relevant entities using relation-specific attention weights. This ensures that directly connected entities and relations are retrieved first, providing a foundation for further exploration. Next, the agent expands the search globally by identifying cross-modal paths through iterative message passing. This allows the agent to explore deeper semantic relationships beyond immediate neighbors, enhancing the richness of the retrieved information.

Furthermore, the framework is a dual-level retrieval framework that integrates graph-structured knowledge with vector representations through a three-phase retrieval process. First, the framework performs semantic decomposition of the input query $q$ to derive local keywords  $q_l$ and global keywords $q_g$. This step captures both fine-grained and high-level semantic information.
Second, the framework executes hybrid graph-vector matching. An optimized vector database aligns $q_l$ with entity attributes while mapping $q_g$ to relational patterns in the knowledge graph $G=(\mathcal{V},\mathcal{E})$. This hybrid approach ensures that both explicit entity attributes and latent relational semantics are considered.

Finally, to enhance retrieval completeness, the framework performs higher-order context expansion. The retrieved subgraph is expanded to include one-hop neighbors of both retrieved nodes and edges:
\begin{equation}
     \mathcal{A}_g = \{v_i \in \mathcal{V} \wedge(v_i\in \mathcal{N}_v \vee v_i \in \mathcal{N}_e)\} 
\end{equation}
where $\mathcal{N}_v$ and $\mathcal{N}_e$ denote the one-hop neighbors of retrieved nodes and edges, respectively. This step ensures that the retrieved subgraph retains structural integrity while capturing broader contextual relationships. The final answer $\mathcal{A}_g$ is generated using $\mathcal{A}_g = LLM(\mathcal{A}_g)$ with a lightweight LLM.

\subsubsection{Web-based Retrieval Agent for Real-Time Information}

The web retrieval component serves as a critical bridge between information retrieval and natural language generation, significantly enhancing the semantic fidelity and factual grounding of generated text. Our work utilizes the Google Serper API. The system acquires knowledge through parameterized API requests to Google's search engine. For an input query $q$, the retrieval process is formalized as:
\begin{equation}
     \mathcal{R} = Google(q;\theta_{search})
\end{equation}
where $\theta_{search}$ specifies search configuration parameters. We adopt the setting that $\theta_{search} = \{num_{results}=k,language=en,type=web\}$. The API returns structured results $\mathcal{A}_w = \{a_i\}_{i=1}^k$, each containing a title, a snippet, a URL, and positional ranking metadata.



The Google Serper framework demonstrates particular efficacy in real-world deployment scenarios through three principal operational modalities, each addressing the critical requirements of modern knowledge-aware systems. First, the real-time fact verification module computes factual validity scores through neural memory interrogation. Second, the attribution-aware generation protocol ensures traceability through dual-phase attention routing. Third, the adaptive query expansion mechanism addresses vocabulary mismatch through differential term weighting.

\subsection{Decision Agent for Multi-answer Refinement}
\textbf{Consistency Voting.} The framework evaluates the semantic agreement among answers $\{\mathcal{A}_v, \mathcal{A}_g, \mathcal{A}_w\}$ generated by vector-based, graph-based, and web-based retrieval systems using ROUGE-L and BLEU metrics. Summaries $\{\mathcal{S}_v, \mathcal{S}_g, \mathcal{S}_w\}$ are first generated for each answer. ROUGE-L measures the overlap of key information using the Longest Common Subsequence (LCS), defined as:
\begin{equation}
    R_L = \frac{LCS(\mathcal{S}_i, \mathcal{S}_j)}{max(|\mathcal{S}_i|, |\mathcal{S}_j|)}
\end{equation}
where the numerator represents the length of the LCS between summaries, while the denominator normalizes the score. This metric emphasizes consistency in retaining critical factual information.

BLEU evaluates the localized precision of n-gram matches between summaries, defined as:
\begin{equation}
    BLEU = exp(\sum_{n=1}^k w_nlog p_n)\cdot min(1, \frac{|\mathcal{S}_j|}{|\mathcal{S}_i|})
\end{equation}
where $p_n$ represents $n$-gram precision, and $w_n$ denotes weight coefficients. This metric excels in detecting precise matches of terminologies or numerical values.

A weighted fusion of $R_L$ and $BLEU$ is then applied to balance macro-level semantic alignment with micro-level detail consistency, measuring the similarity between any two answers. If the pairwise similarity exceeds a predefined threshold, the result is refined using a Lightweight Language Model (LLM) to produce the final answer A. The framework proceeds to expert model refinement if the similarity is below the threshold.

\textbf{Expert Model Refinement.} For conflicting answers, the framework employs LLMs, Multimodal LLMs (MLLMs) or Cot-based language models (Cot-LMs) to synthesize a refined response by integrating multi-source evidence. The LLM or MLLM processes the original query $q$ and the retrieved evidence to generate the final answer $\mathcal{A}$. This step serves as an expert-guidance,ensuring that the final response is both contextually coherent and factually accurate, even when initial answers exhibit discrepancies.

\section{Experiments}

\subsection{Experimental Setup}
\textbf{Dataset.}  We conduct experiments across two multimodal reasoning benchmarks spanning divergent modality configurations, including complex question answering (ScienceQA) and crisis event classification (CrisisMMD).

\textbf{ScienceQA} \cite{lu2022learn}. This dataset is the first large-scale multimodal benchmark for scientific question answering spanning 3 core disciplines (Natural Science, Social Science, and Formal Science). The dataset contains 21,208 carefully curated examples organized hierarchically across 26 topics, 127 categories, and 379 distinct reasoning skills. Each instance combines textual questions with optional visual contexts (diagrams, charts, or photographs), with a balanced split of 12,726 training, 4,214 validation, and 4,268 test samples. Following the evaluation protocol established in LLaVA \cite{liu2023visual}, we report averaged accuracy across all test samples to assess model performance in multimodal understanding and multi-step scientific reasoning. Notably, 34.6\% of test questions require simultaneous processing of both visual and textual information to derive correct answers.

\textbf{CrisisMMD} \cite{alam2018crisismmd}. This dataset presents a challenging multimodal collection for disaster response applications, comprising approximately 35,000 social media posts containing both visual and textual content from real-world crisis events. It features a comprehensive annotation scheme with seven distinct disaster categories and four granular severity levels. Its unique value lies in capturing authentic user-generated content that preserves natural noise patterns and complex cross-modal relationships inherent in crisis communication. These characteristics make it particularly suitable for evaluating zero-shot adaptation models, as successful performance on this benchmark directly correlates with practical deployment capabilities in dynamic emergency scenarios where clean data and explicit modality alignments are typically unavailable.

\textbf{Implementation Details.}
We utilize DeepSeek-R1-70B for dynamic graph construction and optimize LightRAG's hybrid retrieval mechanism through Qwen2.5-7B's parameter adaptation framework, which is consistent with VaLik \cite{liu2025aligning}. During decision refinement, we employ GPT-4o for ScienceQA dataset processing and GPT-4 for CrisisMMD dataset analysis. All multimodal reasoning workflows operate on a single NVIDIA A800-80GB GPU, seamlessly supporting the concurrent execution of graph neural network computations and retrieval-augmented generation tasks through memory-optimized parallelization.

\begin{table*}[ht]
    \centering
    \caption{Top-1 retrieval performance comparison (Accuracy \%) on the ScienceQA Dataset. \#P denotes the number of trainable parameters. Categories include: NAT (Natural Science), SOC (Social Science), LAN (Language Science), TXT (Text Context), IMG (Image Context), NO (No Context), G1-6 (Grades 1-6), and G7-12 (Grades 7-12). The comparisons presented are based on the state-of-the-art zero-shot learning results obtained from the ScienceQA leaderboard\protect\footnotemark.}
    \begin{tabular}{c|l|c|ccc|ccc|cc|c}
    \hline 
         \multirow{2}{*}{\textbf{Learning}} & \multirow{2}{*}{\textbf{Models}}& \multirow{2}{*}{\textbf{\#P}} & \multicolumn{3}{c|}{\textbf{Subject}} & \multicolumn{3}{c|}{\textbf{Context Modality}} & \multicolumn{2}{c|}{\textbf{Grade}} &\multirow{2}{*}{\textbf{Average}} \\
         \cline{4-11}
         ~ &~ &~ & \textbf{NAT} & \textbf{SOC} & \textbf{LAN} & \textbf{TXT} & \textbf{IMG} & \textbf{NO} & \textbf{G1-6} & \textbf{G7-12} & ~ \\
         \hline
         \multirow{1}{*}{Baseline} & Human & -& 90.23 & 84.97 & 87.48 & 89.60 & 87.50 & 88.10 & 91.59 & 82.42 & 88.40\\
         \hline
         \multirow{5}{*}{Zero-shot LLMs} & ChatGPT \cite{yang2023mm} & - & -	&-	&-	&-	&-&	-	&-	&-	&69.41 \\
          ~ & GPT-3 (0-shot) \cite{lu2022learn}& 173B & 75.04&	66.59&	78.00	&74.24&	65.74	&79.58	&76.36	&69.87	&74.04 \\
         ~ & DDCoT (GPT-3) \cite{zheng2023ddcot}& 175B & 78.60&	73.90&	80.45&	77.27	&69.96&	82.93&	80.65	&73.50	&78.09\\
         ~ & CoT GPT-3 + Doc \cite{hsieh2023tool}& 173B	& - &- &- &- &- & - & - & - &79.91 \\
         ~ & DDCoT (ChatGPT) \cite{zheng2023ddcot} & 175B&80.15&76.72&82.82&78.89 &72.53 &85.02&82.86&75.21&80.15\\
         \hline    
         \multirow{5}{*}{Zero-shot VLMs}  & LaVIN-13B \cite{yang2023mm}& - & - &- &- &- &- & - & - & -& 77.54 \\
         ~ & LLaMA-SciTune \cite{horawalavithana2023scitune}& 7B &84.50& 94.15&	82.91&	88.35&	83.64&	88.74&	85.05&	85.60&	86.11\\     
         ~ & LG-VQA (BLIP-2) \cite{ghosal2023language}& - & - &- &- &- &- & - & - & - & 86.32 \\
         ~ & LG-VQA (CLIP) \cite{ghosal2023language}& - & - &- &- &- &- & - & - & - & 87.22 \\
         ~ & LLaMA-SciTune \cite{horawalavithana2023scitune} & 13B &89.30	&95.61	&87.00	&93.08	&86.67	&91.75&	84.37&	91.30	&90.03 \\
         \hline      
         ~ & Vector-based \cite{liu2025aligning}& 7B & 84.54 &74.24 &86.91& 82.74 &72.53 &90.03 &84.51 &80.28& 82.98\\
          Zero-shot & Graph-based \cite{liu2025aligning}& 7B & 84.15& 75.14& 87.64& 82.99& 73.18& 89.69 &84.40& 80.95& 83.16\\
        Single-agent RAG& Web-based &7B &83.79 & 72.89 & 91.82 & 81.09 & 70.55 & 94.01 & 85.98 & 79.30 & 83.59\\
         ~ & GPT-4o \cite{hurst2024gpt} &- &92.72 & 93.48 & 86.09 & 92.67 & 90.88 & 87.60 & 92.91 & 88.00 & 91.16\\
         \hline
         \multirow{1}{*}{Zero-shot} & \multirow{2}{*}{\model} & \multirow{2}{*}{-} & \multirow{2}{*}{\textbf{94.36}} & \multirow{2}{*}{\textbf{90.66}} & \multirow{2}{*}{\textbf{94.91}} & \multirow{2}{*}{\textbf{93.79}} & \multirow{2}{*}{\textbf{89.94}} & \multirow{2}{*}{\textbf{96.03}} & \multirow{2}{*}{\textbf{94.42}} & \multirow{2}{*}{\textbf{92.49}} & \multirow{2}{*}{\textbf{93.73}}\\
          \multirow{1}{*}{Multi-agent RAG} &~ &~ & ~ & ~ & ~ & ~ & ~ & ~& ~ & ~ \\
         \hline
    \end{tabular}
    \label{tab:tab1}
\end{table*}

\begin{table}
    \centering
    \caption{Top-1 retrieval performance comparison (Accuracy \%) on the CrisisMMD Dataset. The -I indicates instruction-tuned variants. \textbf{Bold} denotes the highest value. Task 1 is a binary classification task, while Task 2 and Task 2 Merged are multi-classification tasks. The comparisons are sourced from \cite{liu2025aligning}, which represents the pioneering LLM-based work on the CrisisMMD Dataset.}
    \scalebox{0.90}{
    \begin{tabular}{c|c|ccc|c}
    \hline
       \multirow{2}{*}{\textbf{Method}} & \multirow{2}{*}{\textbf{\#P}} &  \multirow{2}{*}{\textbf{Task 1}} & \multirow{2}{*}{\textbf{Task 2}} & \textbf{Task 2} & \multirow{2}{*}{\textbf{Average}}\\
       ~& ~& ~ & ~& \textbf{Merged} & ~ \\
        \hline
        \multicolumn{6}{l}{\textbf{Single-modal LLMs}}\\
        \hline
        \multirow{3}{*}{LLaMA-2 \cite{touvron2023llama}} & 7B &62.32& 18.32&21.45 & 34.03\\
        ~ & 13B &63.80&21.82&33.15 &39.59\\
        ~ & 70B &63.15&28.87&36.89 &42.97\\
        \cline{2-6}
        \multirow{3}{*}{Qwen2.5 \cite{yang2024qwen2}} &7B &65.04&44.52&45.33 &51.63 \\
        ~ &32B &67.28 & 46.94& 47.07&53.76\\
        ~ & 72B&67.95 & 50.51 & 50.29&56.25\\
        \cline{2-6}
        GPT-4 \cite{achiam2023gpt} &- &66.83&47.25&49.44 &54.51\\
        \hline
        \multicolumn{6}{l}{\textbf{Multimodal VLMs}}\\
        \hline
        \multirow{3}{*}{Qwen2-VL \cite{wang2024qwen2}} & 2B-I & 47.56 & 7.60 & 7.42 & 20.86\\
        ~ & 7B-I & 62.45 & 32.68 & 34.20 & 43.11\\
        ~ & 72B-I & 65.80 & 47.21 & 48.28 & 53.76\\
        \cline{2-6}
        \multirow{3}{*}{LLaVA \cite{liu2023visual}} & 7B & 54.00 & 28.01 & 30.61 & 37.54\\
        ~ & 13B & 60.58 & 20.14 & 23.44 & 34.72\\
        ~ & 34B &56.44 & 25.15 & 25.07 & 35.55\\
        \cline{2-6}
        \cline{2-6}
        CLIP \cite{radford2021learning}& - & 43.36 & 17.88 & 20.79 &27.34\\
        GPT-4o \cite{hurst2024gpt} & -& 68.20 & 47.58 & 49.55 & 55.11\\
        \hline
        \multicolumn{6}{l}{\textbf{Single-agent RAG}}\\
        \hline
        Vector-based \cite{liu2025aligning}&7B & 67.49& 45.11&45.94 & 52.85\\
        Graph-based \cite{liu2025aligning}&7B &68.90&50.02&50.69 & 56.54\\
        \hline
        \multicolumn{6}{l}{\textbf{Multi-agent RAG}}\\
        \hline
        \multirow{2}{*}{HM-RAG} & \multirow{2}{*}{-} & \multirow{2}{*}{\textbf{72.06}} & \multirow{2}{*}{\textbf{51.50}}& \multirow{2}{*}{\textbf{52.09}} & \multirow{2}{*}{\textbf{58.55}}\\  
        ~& ~& ~ & ~& ~& ~ \\
    \hline
    \end{tabular}}
    \label{tab:tab2}
\end{table}

\subsection{Main Results}

In this section, we conduct a systematic evaluation of {\model} against state-of-the-art zero-shot LLMs, VLMs, and RAG-enhanced approaches across multiple benchmarks. The results are presented in Table \ref{tab:tab1} and Table \ref{tab:tab2}, which demonstrate \textbf{the consistent superiority of {\model} over all comparative methods}.

\subsubsection{Results on ScienceQA}
Table \ref{tab:tab1} systematically quantifies the multimodal question-answering performance of {\model} and existing zero-shot approaches
on the ScienceQA dataset.
As shown in the table, HM-RAG establishes the state-of-the-art average accuracy of 93.73\%, surpassing the previous best zero-shot VLM method LLaMA-SciTune and GPT-4o by 4.11\% and 2.82\%, respectively, and significantly outperforming the single-agent RAG variants. Compared to vector-based, graph-based, and web-based baselines, HM-RAG achieves 12.95\%, 12.71\%, and 12.13\% absolute improvements, respectively. Notable gains are observed in the accuracy of Social Science (SOC) tasks, where the improvements over web-based and graph-based baselines reach 24.38\% and 20.65\%, respectively. The framework also exceeds human expert performance by 6.03\%.

\subsubsection{Results on CrisisMMD}
Table \ref{tab:tab2} presents a comprehensive evaluation of multimodal understanding capabilities on the CrisisMMD benchmark. Our analysis reveals three key observations. First, multimodal enhanced LLMs consistently outperform both text-only LLMs and specialized VLMs across all tasks. The proposed method achieves state-of-the-art performance with an average accuracy of 58.55\%, representing 2.44\% and 3.44\% absolute improvements over the strongest baseline (GPT-4o) and text-only variant (Qwen2.5-72B), respectively, despite using only 7B parameters.

Second, the model scale exhibits a non-linear correlation with performance gains. While Qwen2.5-72B (text-only) achieves 56.25\% average accuracy, our 7B multimodal enhanced variant attains an absolute improvement of 2.3\%, demonstrating superior parameter efficiency. This trend holds across modalities, with Qwen2-VL-72B-I (VLM) underperforming our method by 4.79\% despite equivalent parameter counts.

Third, multimodal integration significantly impacts task performance. Our method shows 5.7\% and 2.01\% improvements in average accuracy over its text-only and graph-only variants, respectively, which highlights the effectiveness of multi-source reasoning. Notably, the accuracy of 72.06\% on Task 1 establishes a new benchmark, outperforming GPT-4o by 3.86\% and demonstrating robust visual-textual alignment capabilities.

\begin{figure*}
    \centering
    \includegraphics[width=0.8\linewidth]{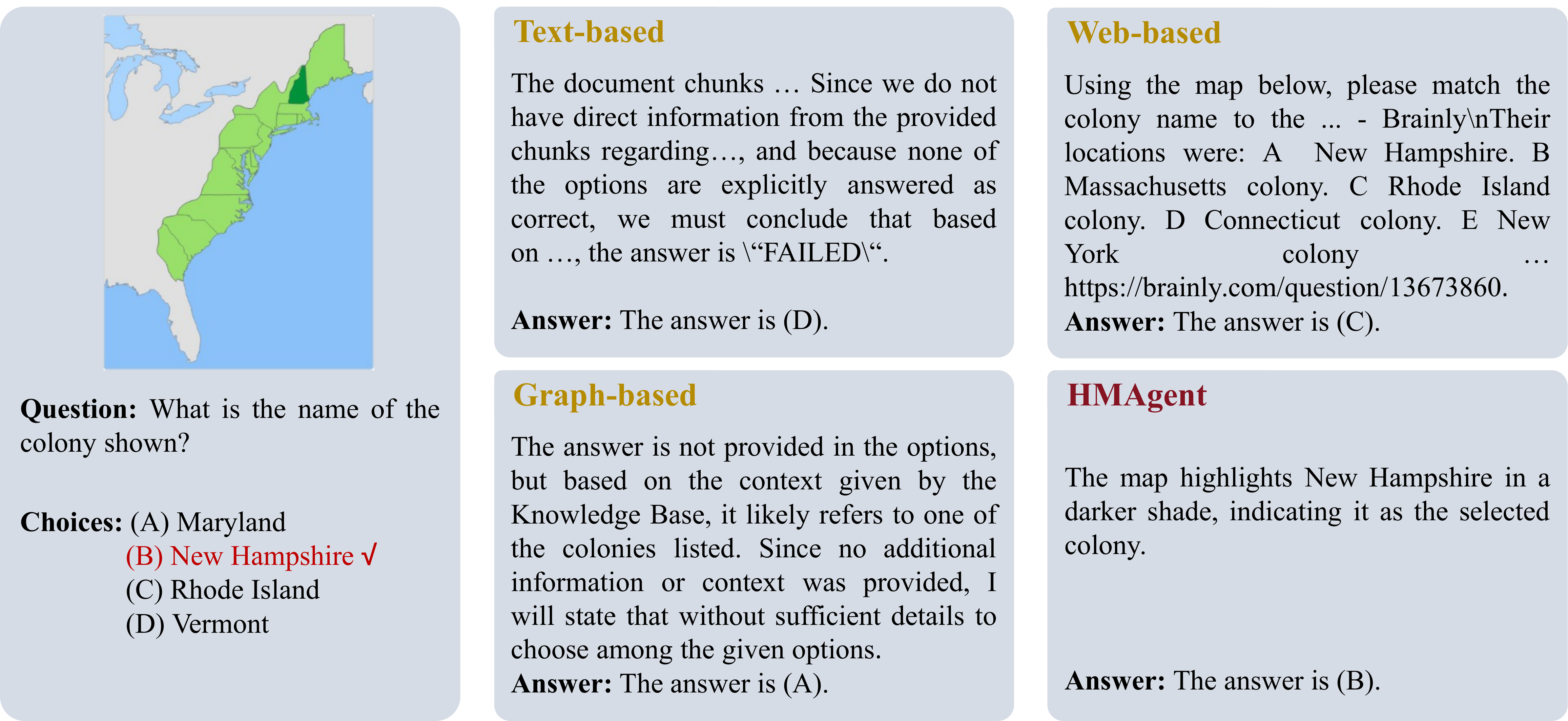}
    \caption{Case Study: Comparison Between {\model} and the Baseline Methods (Vector-based, Graph-based, and Web-based Retrieval Agent).}
    \label{fig:exp1}
\end{figure*}

\begin{table*}[h]
    \centering
    \caption{Performance comparison across different variants of {\model} on the ScienceQA Dataset. Components include: VA (Vector-based Retrieval Agent), GA (Graph-based Retrieval Agent), WA (Web-based Retrieval Agent), and DA (Decision Agent).}
    \begin{tabular}{cccc|ccccccccc}
    \hline
         \multicolumn{4}{c|}{Agent Configuration} & \multirow{2}{*}{NAT} & \multirow{2}{*}{SOC} & \multirow{2}{*}{LAN} & \multirow{2}{*}{TXT} & \multirow{2}{*}{IMG} & \multirow{2}{*}{NO} & \multirow{2}{*}{G1-6} & \multirow{2}{*}{G7-12} & \multirow{2}{*}{Average} \\
         \cline{0-3}
         VA & GA & WA & DA & ~ & ~ & ~ & ~ & ~ & ~ & ~ & ~ & ~ \\
         \hline

         $\times$ & $\checkmark$ & $\checkmark$ & $\checkmark$ &  90.72 & 88.08 & 94.09 & 89.30 & 84.58 & 95.68 & 92.47 & 88.46 & 91.04  \\
         
         $\checkmark$ & $\times$ & $\checkmark$ & $\checkmark$ &  91.21 & 87.96 & 94.73 & 90.32 & 85.62 & 95.61 & 92.22 & 90.05 & 91.44  \\

         $\checkmark$ & $\checkmark$ & $\times$ & $\checkmark$ &  88.99 & 84.81 & 90.27 & 88.17 & 83.09 & 91.78 & 89.46 & 86.62 & 88.45  \\

         $\checkmark$ & $\checkmark$ & $\checkmark$ & $\times$ & 83.79 & 72.89 & 91.82 & 81.09 & 70.55 & 94.01 & 85.98 & 79.30 & 83.59  \\
         $\checkmark$ & $\checkmark$ & $\checkmark$ & $\checkmark$ & \textbf{94.36} & \textbf{90.66} & \textbf{94.91} & \textbf{93.79} & \textbf{89.94} & \textbf{96.03} & \textbf{94.42} & \textbf{92.49} & \textbf{93.73} \\
         \hline
    \end{tabular}
    \label{tab:ab1}
\end{table*}
\footnotetext{\url{https://scienceqa.github.io/leaderboard.html}}
\subsection{Qualitative Analysis}
We provide a prediction example as shown in Figure \ref{fig:exp1} to demonstrate the effectiveness of our proposed model. This example was carefully chosen to showcase the model's ability to handle complex patterns and make accurate choices. For readers interested in additional cases, a more detailed set of examples is provided in Appendix \ref{sec:exps}. In the given example, the multi-source retrieval agents all produce incorrect results since there is no relevant information recorded for this question in the database. To cope with the situation, the expert refinement in the decision agent is used to perform high-level thinking to derive the correct result. This manifests the proficiency of our model in informed decision-making, which assures enhanced robustness compared to relying on a single type of retrieval mechanism.

\subsection{Ablation Studies}
Table \ref{tab:ab1} presents a systematic evaluation of individual agent components' contributions through controlled ablation studies on ScienceQA. Three key insights emerge regarding the framework's design. First, the decision agent (DA) establishes itself as the most critical element, with its removal triggering the most substantial performance decline at 10.82\%. This component proves particularly vital for synthesizing multi-source decisions, as evidenced by significant accuracy reductions of 21.56\% in image-based tasks and 19.60\% in social reasoning tasks when DA is disabled. Second, the web-based retrieval agent (WA) demonstrates robust integration capabilities. Deactivating WA leads to an average performance decrease of 5.63\%, with a more pronounced impact on grade 7-12 tasks, showing a 6.35\% accuracy drop. Third, the fully integrated agent system achieves peak performance at 93.73\%, surpassing the best ablated configuration by a notable margin of 2.44\%. This optimal configuration delivers consistent enhancements across all task categories, particularly excelling in multimodal scenarios with 3.70\% improvement in text-based tasks and 4.80\% in image-based tasks compared to the baselines. The framework also shows superior handling of complex queries, attaining 2.64\% higher accuracy for grade 7-12 problems. These empirical outcomes substantiate the architectural effectiveness in orchestrating specialized agents for holistic multimodal reasoning.



\section{Conclusion}
\label{sec:conclu}

In this paper, we introduced {\model}, a novel Hierarchical Multi-Agent Multimodal Retrieval-Augmented Generation framework designed to address the challenges of complex multimodal query processing and knowledge synthesis. HM-RAG pioneers collaborative intelligence by integrating specialized agents for query decomposition, multi-source retrieval, and decision refinement, enabling dynamic knowledge synthesis across structured, unstructured, and graph-based data. Through extensive experiments on the ScienceQA and CrisisMMD benchmarks, {\model} demonstrated state-of-the-art performance in the accuracy of multimodal question answering and classification, with significant improvements over all categories of baseline methods. 
Our work advances RAG systems by effectively addressing critical challenges in multimodal reasoning and knowledge synthesis, paving the way for more robust and adaptable information retrieval and generation systems in diverse application domains.


\clearpage
\bibliographystyle{ACM-Reference-Format}
\bibliography{main}


\begin{thebibliography}{59}


\ifx \showCODEN    \undefined \def \showCODEN     #1{\unskip}     \fi
\ifx \showISBNx    \undefined \def \showISBNx     #1{\unskip}     \fi
\ifx \showISBNxiii \undefined \def \showISBNxiii  #1{\unskip}     \fi
\ifx \showISSN     \undefined \def \showISSN      #1{\unskip}     \fi
\ifx \showLCCN     \undefined \def \showLCCN      #1{\unskip}     \fi
\ifx \shownote     \undefined \def \shownote      #1{#1}          \fi
\ifx \showarticletitle \undefined \def \showarticletitle #1{#1}   \fi
\ifx \showURL      \undefined \def \showURL       {\relax}        \fi
\providecommand\bibfield[2]{#2}
\providecommand\bibinfo[2]{#2}
\providecommand\natexlab[1]{#1}
\providecommand\showeprint[2][]{arXiv:#2}

\bibitem[Achiam et~al\mbox{.}(2023)]%
        {achiam2023gpt}
\bibfield{author}{\bibinfo{person}{Josh Achiam}, \bibinfo{person}{Steven Adler}, \bibinfo{person}{Sandhini Agarwal}, \bibinfo{person}{Lama Ahmad}, \bibinfo{person}{Ilge Akkaya}, \bibinfo{person}{Florencia~Leoni Aleman}, \bibinfo{person}{Diogo Almeida}, \bibinfo{person}{Janko Altenschmidt}, \bibinfo{person}{Sam Altman}, \bibinfo{person}{Shyamal Anadkat}, {et~al\mbox{.}}} \bibinfo{year}{2023}\natexlab{}.
\newblock \showarticletitle{GPT-4 Technical Report}.
\newblock \bibinfo{journal}{\emph{arXiv preprint arXiv:2303.08774}} (\bibinfo{year}{2023}).
\newblock


\bibitem[Alam et~al\mbox{.}(2018)]%
        {alam2018crisismmd}
\bibfield{author}{\bibinfo{person}{Firoj Alam}, \bibinfo{person}{Ferda Ofli}, {and} \bibinfo{person}{Muhammad Imran}.} \bibinfo{year}{2018}\natexlab{}.
\newblock \showarticletitle{CrisisMMD: Multimodal Twitter Datasets from Natural Disasters}. In \bibinfo{booktitle}{\emph{Proceedings of the International AAAI Conference on Web and Social Media}}, Vol.~\bibinfo{volume}{12}.
\newblock


\bibitem[Anand et~al\mbox{.}(2023)]%
        {anand2023context}
\bibfield{author}{\bibinfo{person}{Abhijit Anand}, \bibinfo{person}{Vinay Setty}, \bibinfo{person}{Avishek Anand}, {et~al\mbox{.}}} \bibinfo{year}{2023}\natexlab{}.
\newblock \showarticletitle{Context Aware Query Rewriting for Text Rankers using LLM}.
\newblock \bibinfo{journal}{\emph{arXiv preprint arXiv:2308.16753}} (\bibinfo{year}{2023}).
\newblock


\bibitem[Asai et~al\mbox{.}(2023)]%
        {asai2023self}
\bibfield{author}{\bibinfo{person}{Akari Asai}, \bibinfo{person}{Zeqiu Wu}, \bibinfo{person}{Yizhong Wang}, \bibinfo{person}{Avirup Sil}, {and} \bibinfo{person}{Hannaneh Hajishirzi}.} \bibinfo{year}{2023}\natexlab{}.
\newblock \showarticletitle{Self-rag: Learning to retrieve, generate, and critique through self-reflection}.
\newblock \bibinfo{journal}{\emph{arXiv preprint arXiv:2310.11511}} (\bibinfo{year}{2023}).
\newblock


\bibitem[Bag et~al\mbox{.}(2024)]%
        {bag2024rag}
\bibfield{author}{\bibinfo{person}{Sukanya Bag}, \bibinfo{person}{Ayushman Gupta}, \bibinfo{person}{Rajat Kaushik}, {and} \bibinfo{person}{Chirag Jain}.} \bibinfo{year}{2024}\natexlab{}.
\newblock \showarticletitle{RAG Beyond Text: Enhancing Image Retrieval in RAG Systems}. In \bibinfo{booktitle}{\emph{2024 International Conference on Electrical, Computer and Energy Technologies (ICECET}}. IEEE, \bibinfo{pages}{1--6}.
\newblock


\bibitem[Bonomo and Bianco(2025)]%
        {bonomo2025visual}
\bibfield{author}{\bibinfo{person}{Mirco Bonomo} {and} \bibinfo{person}{Simone Bianco}.} \bibinfo{year}{2025}\natexlab{}.
\newblock \showarticletitle{Visual RAG: Expanding MLLM Visual Knowledge without Fine-tuning}.
\newblock \bibinfo{journal}{\emph{arXiv preprint arXiv:2501.10834}} (\bibinfo{year}{2025}).
\newblock


\bibitem[Chen et~al\mbox{.}(2024)]%
        {chen2024mllm}
\bibfield{author}{\bibinfo{person}{Zhanpeng Chen}, \bibinfo{person}{Chengjin Xu}, \bibinfo{person}{Yiyan Qi}, {and} \bibinfo{person}{Jian Guo}.} \bibinfo{year}{2024}\natexlab{}.
\newblock \showarticletitle{MLLM Is a Strong Reranker: Advancing Multimodal Retrieval-augmented Generation via Knowledge-enhanced Reranking and Noise-injected Training}.
\newblock \bibinfo{journal}{\emph{arXiv preprint arXiv:2407.21439}} (\bibinfo{year}{2024}).
\newblock


\bibitem[Cheng et~al\mbox{.}(2024)]%
        {cheng2024exploring}
\bibfield{author}{\bibinfo{person}{Yuheng Cheng}, \bibinfo{person}{Ceyao Zhang}, \bibinfo{person}{Zhengwen Zhang}, \bibinfo{person}{Xiangrui Meng}, \bibinfo{person}{Sirui Hong}, \bibinfo{person}{Wenhao Li}, \bibinfo{person}{Zihao Wang}, \bibinfo{person}{Zekai Wang}, \bibinfo{person}{Feng Yin}, \bibinfo{person}{Junhua Zhao}, {et~al\mbox{.}}} \bibinfo{year}{2024}\natexlab{}.
\newblock \showarticletitle{Exploring Large Language Model based Intelligent Agents: Definitions, Methods, and Prospects}.
\newblock \bibinfo{journal}{\emph{arXiv preprint arXiv:2401.03428}} (\bibinfo{year}{2024}).
\newblock


\bibitem[Dong et~al\mbox{.}(2024)]%
        {dong2024advanced}
\bibfield{author}{\bibinfo{person}{Yuxin Dong}, \bibinfo{person}{Shuo Wang}, \bibinfo{person}{Hongye Zheng}, \bibinfo{person}{Jiajing Chen}, \bibinfo{person}{Zhenhong Zhang}, {and} \bibinfo{person}{Chihang Wang}.} \bibinfo{year}{2024}\natexlab{}.
\newblock \showarticletitle{Advanced RAG Models with Graph Structures: Optimizing Complex Knowledge Reasoning and Text Generation}. In \bibinfo{booktitle}{\emph{2024 5th International Symposium on Computer Engineering and Intelligent Communications (ISCEIC)}}. IEEE, \bibinfo{pages}{626--630}.
\newblock


\bibitem[Dosovitskiy et~al\mbox{.}(2020)]%
        {dosovitskiy2020image}
\bibfield{author}{\bibinfo{person}{Alexey Dosovitskiy}, \bibinfo{person}{Lucas Beyer}, \bibinfo{person}{Alexander Kolesnikov}, \bibinfo{person}{Dirk Weissenborn}, \bibinfo{person}{Xiaohua Zhai}, \bibinfo{person}{Thomas Unterthiner}, \bibinfo{person}{Mostafa Dehghani}, \bibinfo{person}{Matthias Minderer}, \bibinfo{person}{Georg Heigold}, \bibinfo{person}{Sylvain Gelly}, {et~al\mbox{.}}} \bibinfo{year}{2020}\natexlab{}.
\newblock \showarticletitle{An Image is Worth 16x16 Words: Transformers for Image Recognition at Scale}.
\newblock \bibinfo{journal}{\emph{arXiv preprint arXiv:2010.11929}} (\bibinfo{year}{2020}).
\newblock


\bibitem[e~Aquino et~al\mbox{.}(2025)]%
        {e2025rag}
\bibfield{author}{\bibinfo{person}{Gustavo de~Aquino e Aquino}, \bibinfo{person}{N{\'a}dila da~Silva de Azevedo}, \bibinfo{person}{Leandro Youiti~Silva Okimoto}, \bibinfo{person}{Leonardo Yuto~Suzuki Camelo}, \bibinfo{person}{Hendrio~Luis de Souza~Bragan{\c{c}}a}, \bibinfo{person}{Rubens Fernandes}, \bibinfo{person}{Andre Printes}, \bibinfo{person}{F{\'a}bio Cardoso}, \bibinfo{person}{Raimundo Gomes}, {and} \bibinfo{person}{Israel~Gondres Torn{\'e}}.} \bibinfo{year}{2025}\natexlab{}.
\newblock \showarticletitle{From RAG to Multi-Agent Systems: A Survey of Modern Approaches in LLM Development}.
\newblock  (\bibinfo{year}{2025}).
\newblock


\bibitem[Edge et~al\mbox{.}(2024)]%
        {edge2024local}
\bibfield{author}{\bibinfo{person}{Darren Edge}, \bibinfo{person}{Ha Trinh}, \bibinfo{person}{Newman Cheng}, \bibinfo{person}{Joshua Bradley}, \bibinfo{person}{Alex Chao}, \bibinfo{person}{Apurva Mody}, \bibinfo{person}{Steven Truitt}, \bibinfo{person}{Dasha Metropolitansky}, \bibinfo{person}{Robert~Osazuwa Ness}, {and} \bibinfo{person}{Jonathan Larson}.} \bibinfo{year}{2024}\natexlab{}.
\newblock \showarticletitle{From Local to Global: A GraphRAG Approach to Query-Focused Summarization}.
\newblock \bibinfo{journal}{\emph{arXiv preprint arXiv:2404.16130}} (\bibinfo{year}{2024}).
\newblock


\bibitem[Faysse et~al\mbox{.}(2024)]%
        {faysse2024colpali}
\bibfield{author}{\bibinfo{person}{Manuel Faysse}, \bibinfo{person}{Hugues Sibille}, \bibinfo{person}{Tony Wu}, \bibinfo{person}{Bilel Omrani}, \bibinfo{person}{Gautier Viaud}, \bibinfo{person}{C{\'e}line Hudelot}, {and} \bibinfo{person}{Pierre Colombo}.} \bibinfo{year}{2024}\natexlab{}.
\newblock \showarticletitle{ColPali: Efficient Document Retrieval with Vision Language Models}. In \bibinfo{booktitle}{\emph{The Thirteenth International Conference on Learning Representations}}.
\newblock


\bibitem[Fey et~al\mbox{.}(2023)]%
        {fey2023relational}
\bibfield{author}{\bibinfo{person}{Matthias Fey}, \bibinfo{person}{Weihua Hu}, \bibinfo{person}{Kexin Huang}, \bibinfo{person}{Jan~Eric Lenssen}, \bibinfo{person}{Rishabh Ranjan}, \bibinfo{person}{Joshua Robinson}, \bibinfo{person}{Rex Ying}, \bibinfo{person}{Jiaxuan You}, {and} \bibinfo{person}{Jure Leskovec}.} \bibinfo{year}{2023}\natexlab{}.
\newblock \showarticletitle{Relational Deep Learning: Graph Representation Learning on Relational Databases}.
\newblock \bibinfo{journal}{\emph{arXiv preprint arXiv:2312.04615}} (\bibinfo{year}{2023}).
\newblock


\bibitem[Gao et~al\mbox{.}(2023)]%
        {gao2023retrieval}
\bibfield{author}{\bibinfo{person}{Yunfan Gao}, \bibinfo{person}{Yun Xiong}, \bibinfo{person}{Xinyu Gao}, \bibinfo{person}{Kangxiang Jia}, \bibinfo{person}{Jinliu Pan}, \bibinfo{person}{Yuxi Bi}, \bibinfo{person}{Yi Dai}, \bibinfo{person}{Jiawei Sun}, \bibinfo{person}{Haofen Wang}, {and} \bibinfo{person}{Haofen Wang}.} \bibinfo{year}{2023}\natexlab{}.
\newblock \showarticletitle{Retrieval-Augmented Generation for Large Language Models: A Survey}.
\newblock \bibinfo{journal}{\emph{arXiv preprint arXiv:2312.10997}}  \bibinfo{volume}{2} (\bibinfo{year}{2023}).
\newblock


\bibitem[Genesis and Keane(2025)]%
        {genesis2025integrating}
\bibfield{author}{\bibinfo{person}{Jeanie Genesis} {and} \bibinfo{person}{Frazier Keane}.} \bibinfo{year}{2025}\natexlab{}.
\newblock \showarticletitle{Integrating Knowledge Retrieval with Generation: A Comprehensive Survey of RAG Models in NLP}.
\newblock  (\bibinfo{year}{2025}).
\newblock


\bibitem[Ghosal et~al\mbox{.}(2023)]%
        {ghosal2023language}
\bibfield{author}{\bibinfo{person}{Deepanway Ghosal}, \bibinfo{person}{Navonil Majumder}, \bibinfo{person}{Roy Ka-Wei Lee}, \bibinfo{person}{Rada Mihalcea}, {and} \bibinfo{person}{Soujanya Poria}.} \bibinfo{year}{2023}\natexlab{}.
\newblock \showarticletitle{Language Guided Visual Question Answering: Elevate Your Multimodal Language Model Using Knowledge-Enriched Prompts}.
\newblock \bibinfo{journal}{\emph{arXiv preprint arXiv:2310.20159}} (\bibinfo{year}{2023}).
\newblock


\bibitem[Guo et~al\mbox{.}(2024)]%
        {guo2024lightrag}
\bibfield{author}{\bibinfo{person}{Zirui Guo}, \bibinfo{person}{Lianghao Xia}, \bibinfo{person}{Yanhua Yu}, \bibinfo{person}{Tu Ao}, {and} \bibinfo{person}{Chao Huang}.} \bibinfo{year}{2024}\natexlab{}.
\newblock \showarticletitle{LightRAG: Simple and Fast Retrieval-Augmented Generation}.
\newblock \bibinfo{journal}{\emph{arXiv preprint arXiv:2410.05779}} (\bibinfo{year}{2024}).
\newblock


\bibitem[Gupta et~al\mbox{.}(2024)]%
        {gupta2024comprehensive}
\bibfield{author}{\bibinfo{person}{Shailja Gupta}, \bibinfo{person}{Rajesh Ranjan}, {and} \bibinfo{person}{Surya~Narayan Singh}.} \bibinfo{year}{2024}\natexlab{}.
\newblock \showarticletitle{A Comprehensive Survey of Retrieval-Augmented Generation (RAG): Evolution, Current Landscape and Future Directions}.
\newblock \bibinfo{journal}{\emph{arXiv preprint arXiv:2410.12837}} (\bibinfo{year}{2024}).
\newblock


\bibitem[Guu et~al\mbox{.}(2020)]%
        {guu2020retrieval}
\bibfield{author}{\bibinfo{person}{Kelvin Guu}, \bibinfo{person}{Kenton Lee}, \bibinfo{person}{Zora Tung}, \bibinfo{person}{Panupong Pasupat}, {and} \bibinfo{person}{Mingwei Chang}.} \bibinfo{year}{2020}\natexlab{}.
\newblock \showarticletitle{Retrieval Augmented Language Model Pre-Training}. In \bibinfo{booktitle}{\emph{International Conference on Machine Learning}}. PMLR, \bibinfo{pages}{3929--3938}.
\newblock


\bibitem[Han et~al\mbox{.}(2025)]%
        {han2025mdocagent}
\bibfield{author}{\bibinfo{person}{Siwei Han}, \bibinfo{person}{Peng Xia}, \bibinfo{person}{Ruiyi Zhang}, \bibinfo{person}{Tong Sun}, \bibinfo{person}{Yun Li}, \bibinfo{person}{Hongtu Zhu}, {and} \bibinfo{person}{Huaxiu Yao}.} \bibinfo{year}{2025}\natexlab{}.
\newblock \showarticletitle{MDocAgent: A Multi-Modal Multi-Agent Framework for Document Understanding}.
\newblock \bibinfo{journal}{\emph{arXiv preprint arXiv:2503.13964}} (\bibinfo{year}{2025}).
\newblock


\bibitem[He et~al\mbox{.}(2016)]%
        {he2016deep}
\bibfield{author}{\bibinfo{person}{Kaiming He}, \bibinfo{person}{Xiangyu Zhang}, \bibinfo{person}{Shaoqing Ren}, {and} \bibinfo{person}{Jian Sun}.} \bibinfo{year}{2016}\natexlab{}.
\newblock \showarticletitle{Deep Residual Learning for Image Recognition}. In \bibinfo{booktitle}{\emph{Proceedings of the IEEE Conference on Computer Vision and Pattern Recognition}}. \bibinfo{pages}{770--778}.
\newblock


\bibitem[Horawalavithana et~al\mbox{.}(2023)]%
        {horawalavithana2023scitune}
\bibfield{author}{\bibinfo{person}{Sameera Horawalavithana}, \bibinfo{person}{Sai Munikoti}, \bibinfo{person}{Ian Stewart}, {and} \bibinfo{person}{Henry Kvinge}.} \bibinfo{year}{2023}\natexlab{}.
\newblock \showarticletitle{SCITUNE: Aligning Large Language Models with Scientific Multimodal Instructions}.
\newblock \bibinfo{journal}{\emph{arXiv preprint arXiv:2307.01139}} (\bibinfo{year}{2023}).
\newblock


\bibitem[Hsieh et~al\mbox{.}(2023)]%
        {hsieh2023tool}
\bibfield{author}{\bibinfo{person}{Cheng-Yu Hsieh}, \bibinfo{person}{Si-An Chen}, \bibinfo{person}{Chun-Liang Li}, \bibinfo{person}{Yasuhisa Fujii}, \bibinfo{person}{Alexander Ratner}, \bibinfo{person}{Chen-Yu Lee}, \bibinfo{person}{Ranjay Krishna}, {and} \bibinfo{person}{Tomas Pfister}.} \bibinfo{year}{2023}\natexlab{}.
\newblock \showarticletitle{Tool Documentation Enables Zero-Shot Tool-Usage with Large Language Models}.
\newblock \bibinfo{journal}{\emph{arXiv preprint arXiv:2308.00675}} (\bibinfo{year}{2023}).
\newblock


\bibitem[Hu et~al\mbox{.}(2024)]%
        {hu2024mplug}
\bibfield{author}{\bibinfo{person}{Anwen Hu}, \bibinfo{person}{Haiyang Xu}, \bibinfo{person}{Jiabo Ye}, \bibinfo{person}{Ming Yan}, \bibinfo{person}{Liang Zhang}, \bibinfo{person}{Bo Zhang}, \bibinfo{person}{Chen Li}, \bibinfo{person}{Ji Zhang}, \bibinfo{person}{Qin Jin}, \bibinfo{person}{Fei Huang}, {et~al\mbox{.}}} \bibinfo{year}{2024}\natexlab{}.
\newblock \showarticletitle{mPLUG-DocOwl 1.5: Unified Structure Learning for OCR-free Document Understanding}.
\newblock \bibinfo{journal}{\emph{arXiv preprint arXiv:2403.12895}} (\bibinfo{year}{2024}).
\newblock


\bibitem[Hurst et~al\mbox{.}(2024)]%
        {hurst2024gpt}
\bibfield{author}{\bibinfo{person}{Aaron Hurst}, \bibinfo{person}{Adam Lerer}, \bibinfo{person}{Adam~P Goucher}, \bibinfo{person}{Adam Perelman}, \bibinfo{person}{Aditya Ramesh}, \bibinfo{person}{Aidan Clark}, \bibinfo{person}{AJ Ostrow}, \bibinfo{person}{Akila Welihinda}, \bibinfo{person}{Alan Hayes}, \bibinfo{person}{Alec Radford}, {et~al\mbox{.}}} \bibinfo{year}{2024}\natexlab{}.
\newblock \showarticletitle{GPT-4o System Card}.
\newblock \bibinfo{journal}{\emph{arXiv preprint arXiv:2410.21276}} (\bibinfo{year}{2024}).
\newblock


\bibitem[Izacard et~al\mbox{.}(2022)]%
        {izacard2022few}
\bibfield{author}{\bibinfo{person}{Gautier Izacard}, \bibinfo{person}{Patrick Lewis}, \bibinfo{person}{Maria Lomeli}, \bibinfo{person}{Lucas Hosseini}, \bibinfo{person}{Fabio Petroni}, \bibinfo{person}{Timo Schick}, \bibinfo{person}{Jane Dwivedi-Yu}, \bibinfo{person}{Armand Joulin}, \bibinfo{person}{Sebastian Riedel}, {and} \bibinfo{person}{Edouard Grave}.} \bibinfo{year}{2022}\natexlab{}.
\newblock \showarticletitle{Few-shot Learning with Retrieval Augmented Language Models}.
\newblock \bibinfo{journal}{\emph{arXiv preprint arXiv:2208.03299}} \bibinfo{volume}{1}, \bibinfo{number}{2} (\bibinfo{year}{2022}), \bibinfo{pages}{4}.
\newblock


\bibitem[Jeong(2024a)]%
        {jeong2024graph}
\bibfield{author}{\bibinfo{person}{Cheonsu Jeong}.} \bibinfo{year}{2024}\natexlab{a}.
\newblock \showarticletitle{A Graph-Agent-Based Approach to Enhancing Knowledge-Based QA with Advanced RAG}.
\newblock \bibinfo{journal}{\emph{Knowledge Management Research}} \bibinfo{volume}{25}, \bibinfo{number}{3} (\bibinfo{year}{2024}), \bibinfo{pages}{99--119}.
\newblock


\bibitem[Jeong(2024b)]%
        {jeong2024study}
\bibfield{author}{\bibinfo{person}{Cheonsu Jeong}.} \bibinfo{year}{2024}\natexlab{b}.
\newblock \showarticletitle{A Study on the Implementation Method of an Agent-Based Advanced RAG System Using Graph}.
\newblock \bibinfo{journal}{\emph{arXiv preprint arXiv:2407.19994}} (\bibinfo{year}{2024}).
\newblock


\bibitem[Jiang et~al\mbox{.}(2023)]%
        {jiang2023active}
\bibfield{author}{\bibinfo{person}{Zhengbao Jiang}, \bibinfo{person}{Frank~F Xu}, \bibinfo{person}{Luyu Gao}, \bibinfo{person}{Zhiqing Sun}, \bibinfo{person}{Qian Liu}, \bibinfo{person}{Jane Dwivedi-Yu}, \bibinfo{person}{Yiming Yang}, \bibinfo{person}{Jamie Callan}, {and} \bibinfo{person}{Graham Neubig}.} \bibinfo{year}{2023}\natexlab{}.
\newblock \showarticletitle{Active Retrieval Augmented Generation}.
\newblock \bibinfo{journal}{\emph{arXiv preprint arXiv:2305.06983}} (\bibinfo{year}{2023}).
\newblock


\bibitem[Khattab and Zaharia(2020)]%
        {khattab2020colbert}
\bibfield{author}{\bibinfo{person}{Omar Khattab} {and} \bibinfo{person}{Matei Zaharia}.} \bibinfo{year}{2020}\natexlab{}.
\newblock \showarticletitle{ColBERT: Efficient and Effective Passage Search via Contextualized Late Interaction over BERT}. In \bibinfo{booktitle}{\emph{Proceedings of the 43rd International ACM SIGIR Conference on Research and Development in Information Retrieval}}. \bibinfo{pages}{39--48}.
\newblock


\bibitem[L{\'a}la et~al\mbox{.}(2023)]%
        {lala2023paperqa}
\bibfield{author}{\bibinfo{person}{Jakub L{\'a}la}, \bibinfo{person}{Odhran O'Donoghue}, \bibinfo{person}{Aleksandar Shtedritski}, \bibinfo{person}{Sam Cox}, \bibinfo{person}{Samuel~G Rodriques}, {and} \bibinfo{person}{Andrew~D White}.} \bibinfo{year}{2023}\natexlab{}.
\newblock \showarticletitle{PaperQA: Retrieval-Augmented Generative Agent for Scientific Research}.
\newblock \bibinfo{journal}{\emph{arXiv preprint arXiv:2312.07559}} (\bibinfo{year}{2023}).
\newblock


\bibitem[Lewis et~al\mbox{.}(2020)]%
        {lewis2020retrieval}
\bibfield{author}{\bibinfo{person}{Patrick Lewis}, \bibinfo{person}{Ethan Perez}, \bibinfo{person}{Aleksandra Piktus}, \bibinfo{person}{Fabio Petroni}, \bibinfo{person}{Vladimir Karpukhin}, \bibinfo{person}{Naman Goyal}, \bibinfo{person}{Heinrich K{\"u}ttler}, \bibinfo{person}{Mike Lewis}, \bibinfo{person}{Wen-tau Yih}, \bibinfo{person}{Tim Rockt{\"a}schel}, {et~al\mbox{.}}} \bibinfo{year}{2020}\natexlab{}.
\newblock \showarticletitle{Retrieval-Augmented Generation for Knowledge-Intensive NLP Tasks}.
\newblock \bibinfo{journal}{\emph{Advances in Neural Information Processing Systems}}  \bibinfo{volume}{33} (\bibinfo{year}{2020}), \bibinfo{pages}{9459--9474}.
\newblock


\bibitem[Li et~al\mbox{.}(2023)]%
        {li2023blip}
\bibfield{author}{\bibinfo{person}{Junnan Li}, \bibinfo{person}{Dongxu Li}, \bibinfo{person}{Silvio Savarese}, {and} \bibinfo{person}{Steven Hoi}.} \bibinfo{year}{2023}\natexlab{}.
\newblock \showarticletitle{BLIP-2: Bootstrapping Language-Image Pre-training with Frozen Image Encoders and Large Language Models}. In \bibinfo{booktitle}{\emph{International Conference on Machine Learning}}. PMLR, \bibinfo{pages}{19730--19742}.
\newblock


\bibitem[Li et~al\mbox{.}(2025)]%
        {li2025topology}
\bibfield{author}{\bibinfo{person}{Weijie Li}, \bibinfo{person}{Jin Wang}, \bibinfo{person}{Liang-Chih Yu}, {and} \bibinfo{person}{Xuejie Zhang}.} \bibinfo{year}{2025}\natexlab{}.
\newblock \showarticletitle{Topology-of-Question-Decomposition: Enhancing Large Language Models with Information Retrieval for Knowledge-Intensive Tasks}. In \bibinfo{booktitle}{\emph{Proceedings of the 31st International Conference on Computational Linguistics}}. \bibinfo{pages}{2814--2833}.
\newblock


\bibitem[Liu et~al\mbox{.}(2023)]%
        {liu2023visual}
\bibfield{author}{\bibinfo{person}{Haotian Liu}, \bibinfo{person}{Chunyuan Li}, \bibinfo{person}{Qingyang Wu}, {and} \bibinfo{person}{Yong~Jae Lee}.} \bibinfo{year}{2023}\natexlab{}.
\newblock \showarticletitle{Visual Instruction Tuning}.
\newblock \bibinfo{journal}{\emph{Advances in Neural Information Processing Systems}}  \bibinfo{volume}{36} (\bibinfo{year}{2023}), \bibinfo{pages}{34892--34916}.
\newblock


\bibitem[Liu et~al\mbox{.}(2025a)]%
        {liu2025aligning}
\bibfield{author}{\bibinfo{person}{Junming Liu}, \bibinfo{person}{Siyuan Meng}, \bibinfo{person}{Yanting Gao}, \bibinfo{person}{Song Mao}, \bibinfo{person}{Pinlong Cai}, \bibinfo{person}{Guohang Yan}, \bibinfo{person}{Yirong Chen}, \bibinfo{person}{Zilin Bian}, \bibinfo{person}{Botian Shi}, {and} \bibinfo{person}{Ding Wang}.} \bibinfo{year}{2025}\natexlab{a}.
\newblock \showarticletitle{Aligning Vision to Language: Text-Free Multimodal Knowledge Graph Construction for Enhanced LLMs Reasoning}.
\newblock \bibinfo{journal}{\emph{arXiv preprint arXiv:2503.12972}} (\bibinfo{year}{2025}).
\newblock


\bibitem[Liu et~al\mbox{.}(2025b)]%
        {liu2025siqa}
\bibfield{author}{\bibinfo{person}{Jiawang Liu}, \bibinfo{person}{Ye Tao}, \bibinfo{person}{Fei Wang}, \bibinfo{person}{Hui Li}, {and} \bibinfo{person}{Xiugong Qin}.} \bibinfo{year}{2025}\natexlab{b}.
\newblock \showarticletitle{SiQA: A Large Multi-Modal Question Answering Model for Structured Images Based on RAG}. In \bibinfo{booktitle}{\emph{ICASSP 2025-2025 IEEE International Conference on Acoustics, Speech and Signal Processing (ICASSP)}}. IEEE, \bibinfo{pages}{1--5}.
\newblock


\bibitem[Liu et~al\mbox{.}(2021)]%
        {liu2021swin}
\bibfield{author}{\bibinfo{person}{Ze Liu}, \bibinfo{person}{Yutong Lin}, \bibinfo{person}{Yue Cao}, \bibinfo{person}{Han Hu}, \bibinfo{person}{Yixuan Wei}, \bibinfo{person}{Zheng Zhang}, \bibinfo{person}{Stephen Lin}, {and} \bibinfo{person}{Baining Guo}.} \bibinfo{year}{2021}\natexlab{}.
\newblock \showarticletitle{Swin Transformer: Hierarchical Vision Transformer Using Shifted Windows}. In \bibinfo{booktitle}{\emph{Proceedings of the IEEE/CVF International Conference on Computer Vision}}. \bibinfo{pages}{10012--10022}.
\newblock


\bibitem[Lu et~al\mbox{.}(2022)]%
        {lu2022learn}
\bibfield{author}{\bibinfo{person}{Pan Lu}, \bibinfo{person}{Swaroop Mishra}, \bibinfo{person}{Tanglin Xia}, \bibinfo{person}{Liang Qiu}, \bibinfo{person}{Kai-Wei Chang}, \bibinfo{person}{Song-Chun Zhu}, \bibinfo{person}{Oyvind Tafjord}, \bibinfo{person}{Peter Clark}, {and} \bibinfo{person}{Ashwin Kalyan}.} \bibinfo{year}{2022}\natexlab{}.
\newblock \showarticletitle{Learn to Explain: Multimodal Reasoning via Thought Chains for Science Question Answering}.
\newblock \bibinfo{journal}{\emph{Advances in Neural Information Processing Systems}}  \bibinfo{volume}{35} (\bibinfo{year}{2022}), \bibinfo{pages}{2507--2521}.
\newblock


\bibitem[Luo et~al\mbox{.}(2024)]%
        {luo2024layoutllm}
\bibfield{author}{\bibinfo{person}{Chuwei Luo}, \bibinfo{person}{Yufan Shen}, \bibinfo{person}{Zhaoqing Zhu}, \bibinfo{person}{Qi Zheng}, \bibinfo{person}{Zhi Yu}, {and} \bibinfo{person}{Cong Yao}.} \bibinfo{year}{2024}\natexlab{}.
\newblock \showarticletitle{LayoutLLM: Layout Instruction Tuning with Large Language Models for Document Understanding}. In \bibinfo{booktitle}{\emph{Proceedings of the IEEE/CVF Conference on Computer Vision and Pattern Recognition}}. \bibinfo{pages}{15630--15640}.
\newblock


\bibitem[Mavromatis and Karypis(2024)]%
        {mavromatis2024gnn}
\bibfield{author}{\bibinfo{person}{Costas Mavromatis} {and} \bibinfo{person}{George Karypis}.} \bibinfo{year}{2024}\natexlab{}.
\newblock \showarticletitle{GNN-RAG: Graph Neural Retrieval for Large Language Model Reasoning}.
\newblock \bibinfo{journal}{\emph{arXiv preprint arXiv:2405.20139}} (\bibinfo{year}{2024}).
\newblock


\bibitem[Naveed et~al\mbox{.}(2023)]%
        {naveed2023comprehensive}
\bibfield{author}{\bibinfo{person}{Humza Naveed}, \bibinfo{person}{Asad~Ullah Khan}, \bibinfo{person}{Shi Qiu}, \bibinfo{person}{Muhammad Saqib}, \bibinfo{person}{Saeed Anwar}, \bibinfo{person}{Muhammad Usman}, \bibinfo{person}{Naveed Akhtar}, \bibinfo{person}{Nick Barnes}, {and} \bibinfo{person}{Ajmal Mian}.} \bibinfo{year}{2023}\natexlab{}.
\newblock \showarticletitle{A Comprehensive Overview of Large Language Models}.
\newblock \bibinfo{journal}{\emph{arXiv preprint arXiv:2307.06435}} (\bibinfo{year}{2023}).
\newblock


\bibitem[Procko and Ochoa(2024)]%
        {procko2024graph}
\bibfield{author}{\bibinfo{person}{Tyler~Thomas Procko} {and} \bibinfo{person}{Omar Ochoa}.} \bibinfo{year}{2024}\natexlab{}.
\newblock \showarticletitle{Graph Retrieval-Augmented Generation for Large Language Models: A Survey}. In \bibinfo{booktitle}{\emph{2024 Conference on AI, Science, Engineering, and Technology (AIxSET)}}. IEEE, \bibinfo{pages}{166--169}.
\newblock


\bibitem[Radford et~al\mbox{.}(2021)]%
        {radford2021learning}
\bibfield{author}{\bibinfo{person}{Alec Radford}, \bibinfo{person}{Jong~Wook Kim}, \bibinfo{person}{Chris Hallacy}, \bibinfo{person}{Aditya Ramesh}, \bibinfo{person}{Gabriel Goh}, \bibinfo{person}{Sandhini Agarwal}, \bibinfo{person}{Girish Sastry}, \bibinfo{person}{Amanda Askell}, \bibinfo{person}{Pamela Mishkin}, \bibinfo{person}{Jack Clark}, {et~al\mbox{.}}} \bibinfo{year}{2021}\natexlab{}.
\newblock \showarticletitle{Learning Transferable Visual Models From Natural Language Supervision}. In \bibinfo{booktitle}{\emph{International Conference on Machine Learning}}. PmLR, \bibinfo{pages}{8748--8763}.
\newblock


\bibitem[Riedler and Langer(2024)]%
        {riedler2024beyond}
\bibfield{author}{\bibinfo{person}{Monica Riedler} {and} \bibinfo{person}{Stefan Langer}.} \bibinfo{year}{2024}\natexlab{}.
\newblock \showarticletitle{Beyond Text: Optimizing RAG with Multimodal Inputs for Industrial Applications}.
\newblock \bibinfo{journal}{\emph{arXiv preprint arXiv:2410.21943}} (\bibinfo{year}{2024}).
\newblock


\bibitem[{\c{S}}akar and Emekci(2025)]%
        {csakar2025maximizing}
\bibfield{author}{\bibinfo{person}{Tolga {\c{S}}akar} {and} \bibinfo{person}{Hakan Emekci}.} \bibinfo{year}{2025}\natexlab{}.
\newblock \showarticletitle{Maximizing RAG efficiency: A comparative analysis of RAG methods}.
\newblock \bibinfo{journal}{\emph{Natural Language Processing}} \bibinfo{volume}{31}, \bibinfo{number}{1} (\bibinfo{year}{2025}), \bibinfo{pages}{1--25}.
\newblock


\bibitem[Schick et~al\mbox{.}(2023)]%
        {schick2023toolformer}
\bibfield{author}{\bibinfo{person}{Timo Schick}, \bibinfo{person}{Jane Dwivedi-Yu}, \bibinfo{person}{Roberto Dess{\`\i}}, \bibinfo{person}{Roberta Raileanu}, \bibinfo{person}{Maria Lomeli}, \bibinfo{person}{Eric Hambro}, \bibinfo{person}{Luke Zettlemoyer}, \bibinfo{person}{Nicola Cancedda}, {and} \bibinfo{person}{Thomas Scialom}.} \bibinfo{year}{2023}\natexlab{}.
\newblock \showarticletitle{Toolformer: Language models can teach themselves to use tools}.
\newblock \bibinfo{journal}{\emph{Advances in Neural Information Processing Systems}}  \bibinfo{volume}{36} (\bibinfo{year}{2023}), \bibinfo{pages}{68539--68551}.
\newblock


\bibitem[Su et~al\mbox{.}(2024)]%
        {su2024dragin}
\bibfield{author}{\bibinfo{person}{Weihang Su}, \bibinfo{person}{Yichen Tang}, \bibinfo{person}{Qingyao Ai}, \bibinfo{person}{Zhijing Wu}, {and} \bibinfo{person}{Yiqun Liu}.} \bibinfo{year}{2024}\natexlab{}.
\newblock \showarticletitle{DRAGIN: Dynamic Retrieval Augmented Generation based on the Real-time Information Needs of Large Language Models}.
\newblock \bibinfo{journal}{\emph{arXiv preprint arXiv:2403.10081}} (\bibinfo{year}{2024}).
\newblock


\bibitem[Toro et~al\mbox{.}(2024)]%
        {toro2024dynamic}
\bibfield{author}{\bibinfo{person}{Sabrina Toro}, \bibinfo{person}{Anna~V Anagnostopoulos}, \bibinfo{person}{Susan~M Bello}, \bibinfo{person}{Kai Blumberg}, \bibinfo{person}{Rhiannon Cameron}, \bibinfo{person}{Leigh Carmody}, \bibinfo{person}{Alexander~D Diehl}, \bibinfo{person}{Damion~M Dooley}, \bibinfo{person}{William~D Duncan}, \bibinfo{person}{Petra Fey}, {et~al\mbox{.}}} \bibinfo{year}{2024}\natexlab{}.
\newblock \showarticletitle{Dynamic Retrieval Augmented Generation of Ontologies using Artificial Intelligence (DRAGON-AI)}.
\newblock \bibinfo{journal}{\emph{Journal of Biomedical Semantics}} \bibinfo{volume}{15}, \bibinfo{number}{1} (\bibinfo{year}{2024}), \bibinfo{pages}{19}.
\newblock


\bibitem[Touvron et~al\mbox{.}(2023)]%
        {touvron2023llama}
\bibfield{author}{\bibinfo{person}{Hugo Touvron}, \bibinfo{person}{Louis Martin}, \bibinfo{person}{Kevin Stone}, \bibinfo{person}{Peter Albert}, \bibinfo{person}{Amjad Almahairi}, \bibinfo{person}{Yasmine Babaei}, \bibinfo{person}{Nikolay Bashlykov}, \bibinfo{person}{Soumya Batra}, \bibinfo{person}{Prajjwal Bhargava}, \bibinfo{person}{Shruti Bhosale}, {et~al\mbox{.}}} \bibinfo{year}{2023}\natexlab{}.
\newblock \showarticletitle{Llama 2: Open Foundation and Fine-Tuned Chat Models}.
\newblock \bibinfo{journal}{\emph{arXiv preprint arXiv:2307.09288}} (\bibinfo{year}{2023}).
\newblock


\bibitem[Wang et~al\mbox{.}(2024)]%
        {wang2024qwen2}
\bibfield{author}{\bibinfo{person}{Peng Wang}, \bibinfo{person}{Shuai Bai}, \bibinfo{person}{Sinan Tan}, \bibinfo{person}{Shijie Wang}, \bibinfo{person}{Zhihao Fan}, \bibinfo{person}{Jinze Bai}, \bibinfo{person}{Keqin Chen}, \bibinfo{person}{Xuejing Liu}, \bibinfo{person}{Jialin Wang}, \bibinfo{person}{Wenbin Ge}, {et~al\mbox{.}}} \bibinfo{year}{2024}\natexlab{}.
\newblock \showarticletitle{Qwen2-VL: Enhancing Vision-Language Model's Perception of the World at Any Resolution}.
\newblock \bibinfo{journal}{\emph{arXiv preprint arXiv:2409.12191}} (\bibinfo{year}{2024}).
\newblock


\bibitem[Wu et~al\mbox{.}(2024)]%
        {wu2024medical}
\bibfield{author}{\bibinfo{person}{Junde Wu}, \bibinfo{person}{Jiayuan Zhu}, \bibinfo{person}{Yunli Qi}, \bibinfo{person}{Jingkun Chen}, \bibinfo{person}{Min Xu}, \bibinfo{person}{Filippo Menolascina}, {and} \bibinfo{person}{Vicente Grau}.} \bibinfo{year}{2024}\natexlab{}.
\newblock \showarticletitle{Medical Graph RAG: Towards Safe Medical Large Language Model via Graph Retrieval-Augmented Generation}.
\newblock \bibinfo{journal}{\emph{arXiv preprint arXiv:2408.04187}} (\bibinfo{year}{2024}).
\newblock


\bibitem[Xia et~al\mbox{.}(2024)]%
        {xia2024mmed}
\bibfield{author}{\bibinfo{person}{Peng Xia}, \bibinfo{person}{Kangyu Zhu}, \bibinfo{person}{Haoran Li}, \bibinfo{person}{Tianze Wang}, \bibinfo{person}{Weijia Shi}, \bibinfo{person}{Sheng Wang}, \bibinfo{person}{Linjun Zhang}, \bibinfo{person}{James Zou}, {and} \bibinfo{person}{Huaxiu Yao}.} \bibinfo{year}{2024}\natexlab{}.
\newblock \showarticletitle{MMed-RAG: Versatile Multimodal RAG System for Medical Vision Language Models}.
\newblock \bibinfo{journal}{\emph{arXiv preprint arXiv:2410.13085}} (\bibinfo{year}{2024}).
\newblock


\bibitem[Yang et~al\mbox{.}(2024)]%
        {yang2024qwen2}
\bibfield{author}{\bibinfo{person}{An Yang}, \bibinfo{person}{Baosong Yang}, \bibinfo{person}{Beichen Zhang}, \bibinfo{person}{Binyuan Hui}, \bibinfo{person}{Bo Zheng}, \bibinfo{person}{Bowen Yu}, \bibinfo{person}{Chengyuan Li}, \bibinfo{person}{Dayiheng Liu}, \bibinfo{person}{Fei Huang}, \bibinfo{person}{Haoran Wei}, {et~al\mbox{.}}} \bibinfo{year}{2024}\natexlab{}.
\newblock \showarticletitle{Qwen2.5 Technical Report}.
\newblock \bibinfo{journal}{\emph{arXiv preprint arXiv:2412.15115}} (\bibinfo{year}{2024}).
\newblock


\bibitem[Yang et~al\mbox{.}(2023)]%
        {yang2023mm}
\bibfield{author}{\bibinfo{person}{Xiaocui Yang}, \bibinfo{person}{Wenfang Wu}, \bibinfo{person}{Shi Feng}, \bibinfo{person}{Ming Wang}, \bibinfo{person}{Daling Wang}, \bibinfo{person}{Yang Li}, \bibinfo{person}{Qi Sun}, \bibinfo{person}{Yifei Zhang}, \bibinfo{person}{Xiaoming Fu}, {and} \bibinfo{person}{Soujanya Poria}.} \bibinfo{year}{2023}\natexlab{}.
\newblock \showarticletitle{MM-BigBench: Evaluating Multimodal Models on Multimodal Content Comprehension Tasks}.
\newblock \bibinfo{journal}{\emph{arXiv preprint arXiv:2310.09036}} (\bibinfo{year}{2023}).
\newblock


\bibitem[Zhang et~al\mbox{.}(2024)]%
        {zhang2024raft}
\bibfield{author}{\bibinfo{person}{Tianjun Zhang}, \bibinfo{person}{Shishir~G Patil}, \bibinfo{person}{Naman Jain}, \bibinfo{person}{Sheng Shen}, \bibinfo{person}{Matei Zaharia}, \bibinfo{person}{Ion Stoica}, {and} \bibinfo{person}{Joseph~E Gonzalez}.} \bibinfo{year}{2024}\natexlab{}.
\newblock \showarticletitle{RAFT: Adapting Language Model to Domain Specific RAG}. In \bibinfo{booktitle}{\emph{First Conference on Language Modeling}}.
\newblock


\bibitem[Zheng et~al\mbox{.}(2023)]%
        {zheng2023ddcot}
\bibfield{author}{\bibinfo{person}{Ge Zheng}, \bibinfo{person}{Bin Yang}, \bibinfo{person}{Jiajin Tang}, \bibinfo{person}{Hong-Yu Zhou}, {and} \bibinfo{person}{Sibei Yang}.} \bibinfo{year}{2023}\natexlab{}.
\newblock \showarticletitle{DDCoT: Duty-Distinct Chain-of-Thought Prompting for Multimodal Reasoning in Language Models}.
\newblock \bibinfo{journal}{\emph{Advances in Neural Information Processing Systems}}  \bibinfo{volume}{36} (\bibinfo{year}{2023}), \bibinfo{pages}{5168--5191}.
\newblock


\bibitem[Zhong and Mottin(2023)]%
        {zhong2023knowledge}
\bibfield{author}{\bibinfo{person}{Zhiqiang Zhong} {and} \bibinfo{person}{Davide Mottin}.} \bibinfo{year}{2023}\natexlab{}.
\newblock \showarticletitle{Knowledge-augmented Graph Machine Learning for Drug Discovery: From Precision to Interpretability}. In \bibinfo{booktitle}{\emph{Proceedings of the 29th ACM SIGKDD Conference on Knowledge Discovery and Data Mining}}. \bibinfo{pages}{5841--5842}.
\newblock


\end{thebibliography}

\appendix
\clearpage

\section{Predicted Examples}
\label{sec:exps}

We present additional predicted examples in Figure \ref{fig:exp1}. Furthermore, we include two representative questions with or without image context to assess the models' language reasoning capabilities. {\model} consistently produces accurate answers.

\begin{figure*}[b]
    \centering
    \includegraphics[width=1.0\linewidth]{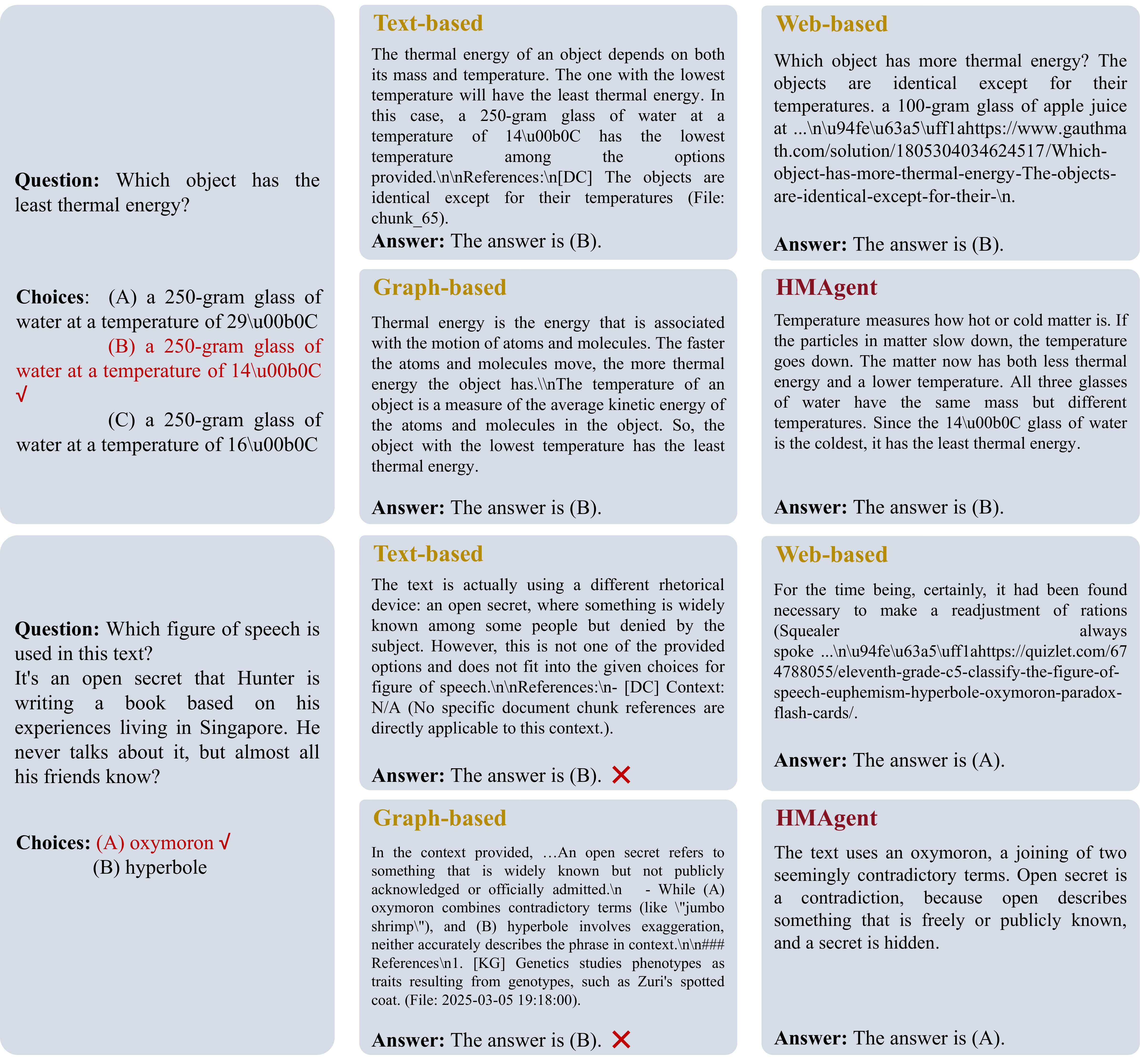}
    \caption{Comparison on single-modal question answering.}
    \label{fig:enter-label}
\end{figure*}

\begin{figure*}[h]
    \centering
    \includegraphics[width=1.0\linewidth]{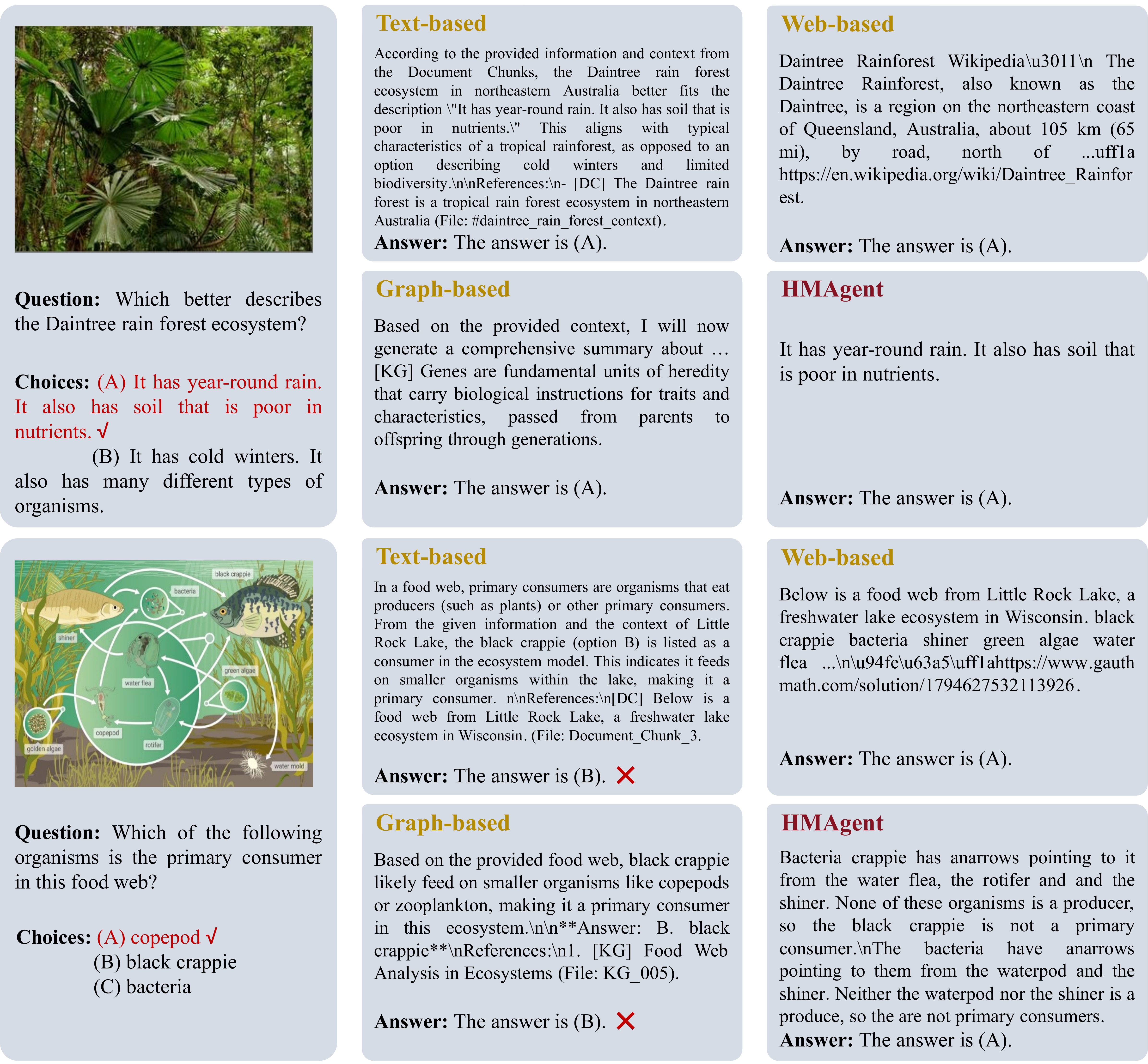}
    \caption{Comparison on multimodal question answering.}
    \label{fig:enter-label}
\end{figure*}

\end{document}